\documentclass[journal,twoside,web]{ieeecolor}
\usepackage{tmi}
\usepackage{cite}
\usepackage{amsmath,amssymb,amsfonts}
\usepackage{algorithmic}
\usepackage{graphicx}
\usepackage{textcomp}
\usepackage{bm}
\usepackage{booktabs}
\usepackage{multirow}
\usepackage{balance}

\usepackage{everypage}
\usepackage{tikz}
\usepackage{lipsum}

\AddEverypageHook{%
 \begin{tikzpicture}[remember picture,overlay, scale=0.6]
 \node[
    scale=0.63,
    opacity=1,
    xshift=0,
    yshift=0,
    color=black
    ]
    at (14, 3.5)  
    {\parbox{24cm}{%
    \begin{tabular}{c}
      This is the author's version of an article that has been published in this journal. Changes were made to this version by the publisher prior to publication. \\
      The final version of record is available at \textcolor{blue}{http://dx.doi.org/10.1109/TMI.2020.3002244}
    \end{tabular}}};
\end{tikzpicture}
}

\AddEverypageHook{%
 \begin{tikzpicture}[remember picture,overlay, scale=0.6]
 \node[
    scale=0.63,
    opacity=1,
    xshift=0,
    yshift=0,
    color=black
    ]
    at (13.5, -41.9)  
    {Copyright (c) 2020 IEEE. Personal use is permitted. For any other purposes, permission must be obtained from the IEEE by emailing pubs-permissions@ieee.org.};
\end{tikzpicture}
}

\def\BibTeX{{\rm B\kern-.05em{\sc i\kern-.025em b}\kern-.08em
    T\kern-.1667em\lower.7ex\hbox{E}\kern-.125emX}}
\begin{document}
\title{Weakly Supervised Deep Nuclei Segmentation Using Partial Points Annotation in Histopathology Images}
\author{Hui Qu, Pengxiang Wu, Qiaoying Huang, Jingru Yi, Zhennan Yan, Kang Li, Gregory M. Riedlinger, Subhajyoti De, Shaoting Zhang, and Dimitris N. Metaxas, \IEEEmembership{Fellow, IEEE}
\thanks{Manuscript received October 25, 2019, revised April 16, 2020, accepted June 4, 2020. This work was supported by NSF-1763523, NSF-1747778, NSF-1733843, NSF-1703883.}
\thanks{H. Qu, P. Wu, Q. Huang, J. Yi and D. N. Metaxas are with the Department
	of Computer Science, Rutgers University, Piscataway, NJ 08854 USA (e-mail: \{hq43, pw241, qh55, jy486, dnm\}@cs.rutgers.edu).}
\thanks{Z. Yan is with SenseBrain Technology Limited LLC, Princeton, NJ 08540 USA (e-mail: zhennanyan@sensebrain.site).}
\thanks{K. Li is with the Department of Orthopaedics, Rutgers New Jersey Medical School, Newark, NJ 07103 USA (e-mail: kl419@rutgers.edu).}
\thanks{S. Zhang is with SenseTime Research, Shanghai 200233 China (e-mail: zhangshaoting@sensetime.com).}
\thanks{G. M. Riedlinger and S. De are with Rutgers Cancer Institute of New Jersey, New Brunswick, NJ 08901 USA (e-mail: \{gr338, sd948\}@cinj.rutgers.edu).}
}

\maketitle

\begin{abstract}
Nuclei segmentation is a fundamental task in histopathology image analysis. Typically, such segmentation tasks require significant effort to manually generate accurate pixel-wise annotations for fully supervised training. To alleviate such tedious and manual effort, in this paper we propose a novel weakly supervised segmentation framework based on partial points annotation, i.e., only a small portion of nuclei locations in each image are labeled.
The framework consists of two learning stages. In the first stage, we design a semi-supervised strategy to learn a detection model from partially labeled nuclei locations. Specifically, an extended Gaussian mask is designed to train an initial model with partially labeled data. Then, self-training with background propagation is proposed to make use of the unlabeled regions to boost nuclei detection and suppress false positives. In the second stage, a segmentation model is trained from the detected nuclei locations in a weakly-supervised fashion. Two types of coarse labels with complementary information are derived from the detected points and are then utilized to train a deep neural network. The fully-connected conditional random field loss is utilized in training to further refine the model without introducing extra computational complexity during inference. The proposed method is extensively evaluated on two nuclei segmentation datasets. The experimental results demonstrate that our method can achieve competitive performance compared to the fully supervised counterpart and the state-of-the-art methods while requiring significantly less annotation effort.
\end{abstract}


\section{Introduction}
\label{sec:introduction}

\IEEEPARstart{H}{istopathology} plays a vital role in cancer diagnosis, prognosis, and treatment decisions. Histopathology slides are created from formalin-fixed paraffin-embedded (FFPE) tissue containing both tumor and surrounding normal tissue. These slides are then stained with agents such as hematoxylin and eosin (H\&E) and immunohistochemical stains that permit the pathologist to ascertain important features. For instance, pathologists routinely determine the type of cancer, stage of cancer, cancer's grade, presence of infiltrating immune cells, and potential treatment options based on histopathology slides. Whole slide imaging allows the pathologist to view the slides digitally as opposed to what was traditionally viewed under a microscope. 
With  improvements in computational power and image analysis algorithms, computational methods~\cite{gurcan2009histopathological,janowczyk2016deep,xing2016robust,zhang2014towards} have been developed for the quantitative and objective analyses of histopathology images, which can reduce the intensive labor and improve the efficiency for pathologists compared with manual examinations. Nuclei segmentation is a critical step in the automatic analysis of histopathology images, because the nuclear features such as average size, density and nucleus-to-cytoplasm ratio are related to the clinical diagnosis and management of cancer. Besides, clinical sequencing of cancer specimens is becoming routine and nuclei segmentation algorithms will play a key role in the proper interpretation of these sequencing results.

Traditional nuclei segmentation algorithms~\cite{veta2013automatic,al2010improved} utilize techniques such as watershed segmentation, physics-based deformable models, level sets and graph cuts. They are designed for a certain type of histopathology images, and cannot work well when there are large variations in tissue types, colors, and nuclear appearances. Learning-based approaches can achieve better performance when dealing with the above variations, as well as some challenging cases like separating touching nuclei. Early learning-based methods, for example Kong et al.~\cite{kong2011partitioning}, trained models using handcrafted features such as color, texture, and other image-level features to segment nuclear regions. Zhang et al.~\cite{zhang2015high} developed a robust segmentation method to delineate cells accurately using Gaussian-based hierarchical voting and repulsive balloon model. Modern deep learning based algorithms~\cite{xing2016automatic,kumar2017dataset,naylor2017nuclei,mahmood2019deep,naylor2018segmentation,raza2019micro,hou2019robust,li2019accurate,qu2019improving,qu2020nuclei} focus on training deep convolutional neural networks (CNNs) for segmentation, and are more effective than previous methods. However, the fully supervised learning of deep neural networks in these methods requires a large amount of training data, which are pixel-wise annotated. It is difficult to collect such  datasets because assigning a nucleus/background class label to every pixel in the image is very time-consuming and requires expert domain knowledge.

In this paper, we describe a novel weakly supervised nuclei segmentation framework for histopathology images using only a portion of annotated nuclear locations. It achieves comparable performance as the fully-supervised methods and about $60\times$ speed-up (10\% points) in the annotation time. Our method consists of two stages: (1) semi-supervised nuclei detection and (2) weakly supervised nuclei segmentation. The goal of the first stage is to train a detector from partial points annotation to predict the locations of all nuclei in training images. A challenge is that there is no clear background information because only  part of the nuclei are labeled in an image. To obtain a good initial detector, we first design an extended Gaussian mask to supervise the training with the labeled nuclear locations and ignore most unlabeled areas. Then, we propose a self-training strategy to make use of the unlabeled areas in images, which refines the background information in an iterative fashion and suppresses false positives.

The detection stage produces central points of all detected nuclei. However, these detected points cannot be directly used to supervise the training of a segmentation model in the second stage. To address this problem, we take advantage of the original image and the shape prior of the nuclei to derive two types of coarse labels from the nuclei points using the Voronoi diagram and the $k$-means clustering algorithm. The coarse labels are used to train a deep convolutional neural network for the segmentation task. A common problem in most weakly supervised segmentation tasks is inaccurate object boundaries due to missing information. Therefore, post-processing like the dense conditional random field (CRF)~\cite{chen2014semantic} or graph search~\cite{yang2018boxnet} is needed to refine the object boundaries, at the expense of extra processing time. Inspired by Tang et al.'s work~\cite{tang2018regularized}, we utilize the dense CRF in the loss function to fine-tune the trained model rather than a post-processing step. This efficient inference is more effective in nuclei segmentation from large Whole Slide Images (WSIs).

This paper is an extension of our previous work~\cite{qu2019weakly} with the following contributions:
\begin{itemize}
	\item A unified framework for nuclei segmentation using only a small portion of nuclear locations (e.g., 10\%).
	\item A novel self-training strategy is proposed in semi-supervised nuclei detection.
	\item More extensive analyses of learning strategies in weakly supervised segmentation.
\end{itemize}

\section{Related Work}
In this section, we provide an overview of the related work in two aspects: (1) deep learning-based nuclei detection and segmentation methods in histopathology images and (2) weakly supervised algorithms in natural and medical images.

\begin{figure*}
	\centering
	\includegraphics[width=\textwidth]{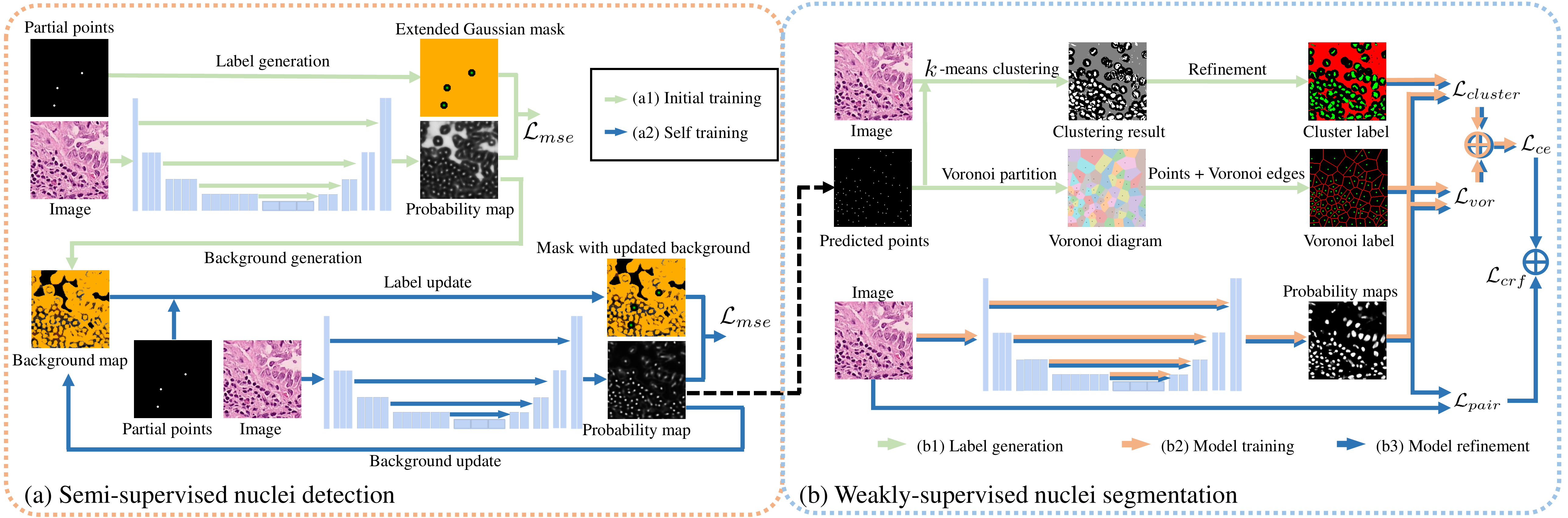}
	\caption{Overview of the proposed method. (a) Semi-supervised nuclei detection module: (b1) initial training step in which an initial detector is trained using the partial points annotation, and (b2) self-training step using background propagation. (b) Weakly supervised nuclei segmentation module: (b1) label generation in which the Voronoi label and cluster label are generated using the detected points and original image, (b2) model training using the cross entropy loss, and (b3) model refinement using the CRF loss.}
	\label{fig:overview}
\end{figure*}

\subsection{Nuclei detection and segmentation using deep learning}
\subsubsection{Nuclei detection}
Deep CNNs have been applied in nuclei detection in recent years. Ciresan et al.~\cite{cirecsan2013mitosis} proposed a mitosis detection method by classifying each pixel using a patch centered on it. Xie et al.~\cite{xie2015beyond} developed a structured regression CNN model (SR-CNN) to utilize topological structure information. Sirinukunwattana et al.~\cite{sirinukunwattana2016locality} improved SR-CNN by using a spatial-constrained layer. Xu et al.~\cite{xu2015stacked} proposed a stacked sparse autoencoder to learn high-level structure information from unlabeled image patches and trained a classifier using the extracted features of the encoder. 
These methods depend on patch-based classification or regression, and thus are computationally expensive for large microscopy images. Later on, the fully convolutional neural network (FCN)~\cite{long2015fully} and its variants were applied in the nuclei detection, which can significantly improve the efficiency in inference by eliminating repeated computations for overlapping patches in the patch-based approaches. A structured regression model for nuclei detection was developed in~\cite{xie2018efficient} to produce better distinctive peaks at cell centroids. Zhou et al.~\cite{zhou2018sfcn} presented a Sibling FCN which detects nuclei and classifies them into sub-categories simultaneously. It takes advantage of the mutual information of both tasks to improve performance. Aside from these supervised training methods, Li et al.~\cite{li2019signet} proposed a semi-supervised learning framework for signet ring cell detection to cope with incomplete annotation and make use of unlabeled images. What is more challenging in our case is that we only have a small portion of nuclei annotated.

\subsubsection{Nuclei segmentation}
Similar to nuclei detection, current deep-learning-based methods can be roughly divided into patch-based and FCN-based categories. In the patch-based category, Su et al.~\cite{su2015robust} utilized sparse denoising autoencoder to segment nuclei. Xing et al.~\cite{xing2016automatic} obtained initial shape probability maps of nuclei by CNN and then incorporated a top-down shape prior model and a bottom-up deformable model for segmentation. Kumar et al.~\cite{kumar2017dataset} formulated the problem as a three-class segmentation task and performed region growing as post-processing based on the initial segmentation results. In the FCN-based category, Naylor et al.~\cite{naylor2018segmentation} solved the problem as a regression task of estimating the nuclei distance map which is beneficial to separate touching or overlapping nuclei. Qu et al.~\cite{qu2019joint} combined the tasks of nuclei segmentation and fine-grained classification into one framework. To solve the problem of insufficient training data, Mahmood et al.~\cite{mahmood2019deep} synthesized additional training images using CycleGAN~\cite{zhu2017unpaired}. And Hou et al.~\cite{hou2019robust} generated images of different tissue types and adopted an importance sampling loss during segmentation according to the quality of synthesized images. We also use the FCN-based framework, but with weak labels (central points).

\subsection{Weakly supervised image segmentation using deep learning}
Compared to fully supervised methods, weakly supervised approaches have the advantage of reducing manual annotation effort. In natural image segmentation, Papandreou et al.~\cite{papandreou2015weakly} proposed Expectation-Maximization (EM) method for training with image-level or bounding-box annotation. Pathak et al.~\cite{pathak2015constrained} added a set of linear constraints on the output space in loss function to exploit the information from image-level labels. Compared to image-level annotation, points annotation has better location information for each object. Bearman et al.~\cite{bearman2016s} incorporated an objectness prior in the loss to guide the training of a CNN, which helps separate objects from background. Scribbles annotation, which requires at least one scribble for every object, is a more informative type of weak label. Lin et al.~\cite{lin2016scribblesup} adopted scribbles annotation to train a graphical model that propagates the information from the scribbles to the unmarked pixels. The most widely used weak annotation is the bounding box, both in natural images~\cite{dai2015boxsup,rajchl2017deepcut} and in medical images~\cite{yang2018boxnet,zhao2018deep}. Kervadec et al.~\cite{kervadec2019constrained} used a small fraction of full labels and imposed a size constraint in their loss function, which achieved good performance but is not applicable for multiple objects of the same class.

Although existing weakly supervised methods have achieved good performance in natural and medical image segmentation, most weak annotations are not suitable for nuclei segmentation task. Image-level annotation cannot be used in medical image segmentation where object classes in images are usually fixed (e.g., nuclei and background in our task). Scribbles annotation is also not suitable for our task due to the small size and large number of nuclei. It is difficult and time-consuming to label an image using bounding boxes for hundreds of nuclei, especially when the density is high. Points annotation is a good choice in terms of preserving information and saving annotation effort, but the objectiveness prior in the points supervision work~\cite{bearman2016s} is not working here since nuclei are small and thus the prior is inaccurate. Different from existing weakly supervised methods, we propose to employ partial points annotation for nuclei segmentation.

\section{Detection with Partial Points}\label{sec:detection}

In this section, we describe the semi-supervised nuclei detection algorithm (Fig.~\ref{fig:overview}(a)), which consists of two steps: initial training with extended Gaussian masks and self-training with background propagation.

\subsection{Initial training with extended Gaussian masks}
The first step of our detection method aims to train an initial detector using the labeled nuclei in each image. However, the points indicating nuclear locations cannot be directly applied for training. They are often used to generate binary masks for pixel classification~\cite{zhou2018sfcn}, or structured proximity masks for regression~\cite{xie2018efficient}. In our case, it is not possible to follow these methods because most areas in an image are unlabeled. In order to tackle this issue, we define an extended Gaussian mask $M$ according to the labeled points:
\begin{equation}\label{eqn:1}
M_i = 
\begin{cases}
\exp\left(-\frac{D_i^2}{2\sigma^2}\right)  & \text{if $D_i < r_1$,} \\
0  &  \text{if $r_1 < D_i < r_2$,} \\
-1 &  \text{otherwise,}	\end{cases}	
\end{equation}
where $D_i$ is the distance from pixel $i$ to the closest labeled point, $r_1$ is average radius of the nuclei and can be calculated using the validation set, $r_2$ is a parameter to control the range of background area and is set to $r_2=2r_1$. $\sigma$ is the Gaussian bandwidth. In $M$, $0$ means the background pixel and $-1$ means unlabeled pixel which will be ignored during training. The underlying assumption is that pixels in the annular area $r_1 < D_i < r_2$ belong to the background, which is reasonable because most nuclei are surrounded by background pixels.

With the extended Gaussian masks, we are able to train a regression model for nuclei detection. We replace the encoder part of U-net~\cite{ronneberger2015u} with the convolution layers of ResNet-34~\cite{he2016deep} (shown in Fig.~\ref{fig:overview}), which is more powerful in representation ability and can be initialized with pretrained parameters. The network is trained with a mean squared loss $\mathcal{L}_{mse}$ with respect to the corresponding extended Gaussian mask:
\begin{equation}\label{eqn:loss}
\mathcal{L}_{mse} = \frac{1}{|\Omega|}\sum_{i\in\Omega}w_i\left(p_i - M_i\right)^2,
\end{equation}
where $\Omega$ is the set consisting of non-ignored pixels, $p_i$ is the predicted probability of being nucleus by the network, and $w_i$ is the weight of pixel $i$. Considering the imbalance between the labeled points and background pixels, we set $w_i=10$ for pixels with mask value greater than $0$ and $w_i=1$ for background pixels. The detection results are obtained by thresholding the probability map and finding the centroids of connected components.

\begin{figure}[t]
	\centering
	\includegraphics[width=0.24\linewidth]{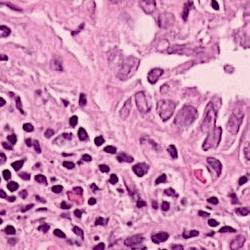} \includegraphics[width=0.24\linewidth]{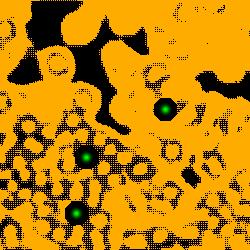}  
	\includegraphics[width=0.24\linewidth]{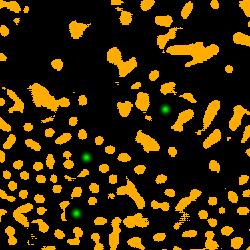}  
	\includegraphics[width=0.24\linewidth]{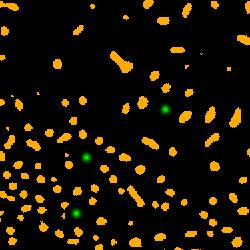} \\
	\begin{minipage} {0.24\linewidth}\centering (a) image \end{minipage} 	
	\begin{minipage} {0.24\linewidth}\centering (b) round 1 \end{minipage}
	\begin{minipage}{0.24\linewidth}\centering	(c) round 2 \end{minipage} 
	\begin{minipage}{0.24\linewidth}\centering	(d) round 3 \end{minipage} 
	\caption{Training masks. The orange color indicates ignored pixels, black is background and green is Gaussian masks centered at labeled points. (a) image, (b)-(d) masks used for training in round 1, 2 3 of self-training, respectively.}
	\label{fig:d2}
\end{figure}

\subsection{Self-training with background propagation}
The detection performance of the initial model is not good enough because of the small number of labeled nuclei and large ignored areas. The unlabeled regions in images can be utilized to improve the performance by semi-supervised learning methods such as self-training. Intuitively, the initial model could predict the nuclei locations on the unlabeled regions. The predicted nuclei are then used to supervise the model training along with the originally labeled nuclei, like what the authors did on cell detection in~\cite{li2019signet}. However, we find that there are too many false positives among the newly detected nuclei because the trained model with the small number of labeled nuclei is not good. The false positives mislead the training during iterations, resulting in worse detection performance. Therefore, we propose an iterative learning strategy to refine the background map during self-training. Because the background tissue and blank areas are easier to be distinguished, producing less false positives.

The process of self-training is shown in Fig.~\ref{fig:overview}(a). In each round of self-training, the background map is firstly obtained from the trained model of the previous round (or the initial model for the first round). Then the background map is combined with the original labeled points to generate a new mask for training. In background map generation, we select background pixels in the probability map if $p_i < 0.1$ or $p_i > 0.7$. The first criterion is straightforward since $p_i$ is the probability of being nuclei. The second one is considered because many background pixels get predicted values close to 1, especially in the first stage. This behavior is expected because most background pixels are ignored during training. If the initialized model predicts them as nuclei pixels, the predictions will remain unchanged. In order to prevent from adding true positive into background, for $p_i > 0.7$ case we only take into account the large connected components of areas greater than the average nuclei area, i.e., $\pi r_1^2$. The updated mask $\tilde{M}$ is finally generated by adding the new background information to the original extended Gaussian mask (See Eqn.~\ref{eqn:mask2}).
\begin{equation}\label{eqn:mask2}
\tilde{M}_i = 
\begin{cases}
M_i & \text{if $D_i < r_2$,} \\
0  &  \text{if $D_i > r_2$ and ($p_i < 0.1$ or $p_i > 0.7$) } \\
-1 &  \text{otherwise,}	\end{cases}	
\end{equation}
An example is shown in Fig.~\ref{fig:d2} to illustrate the masks in different rounds of self-training. The foreground nuclei annotation (green pixels) is kept unchanged during the iterations while the background area (pixels in black) grows up gradually. In the third round, the background has high accuracy and the ignored pixels (orange) are almost all nuclei.

\section{Segmentation with Points}\label{sec:segmentation}
After obtaining a good detection model, we can predict the nuclei locations on all training images. Although the detected points are not 100\% accurate, we have much more information for segmentation compared to the initial partial points. In this section, we describe our weakly supervised method to segment nuclei from the detected points. In particular, our point-level supervision for training a nuclei segmentation model consists of three steps: (1) coarse pixel-level labels generation using the detected points from the nuclei detection stage; (2) segmentation network training with the generated coarse labels; (3) model refinement using the dense CRF loss.

\subsection{From point-level to pixel-level labels}
The point-level labels (detected points) cannot be used directly for the training of a CNN with the cross entropy loss due to the lack of (negative) background labels since all annotated points belong to the (positive) nuclei category. To solve this issue, the first step is to exploit the information we have to generate useful pixel-level labels for both classes. We have the following observations: 
\begin{itemize}
	\item  Each point is expected to be located or close to the center of a nucleus, and the shapes of most nuclei are nearly ellipses, i.e., they are convex.
	\item The colors of nuclei pixels are often different from the surrounding background pixels.
\end{itemize} 
Based on these observations, we propose to utilize the Voronoi diagram and $k$-means clustering methods to produce two types of pixel-level labels.

\subsubsection{Voronoi labels}
\begin{figure}[t]
	\centering
	\begin{minipage}{0.24\linewidth}
		\centering
		\includegraphics[width=\textwidth]{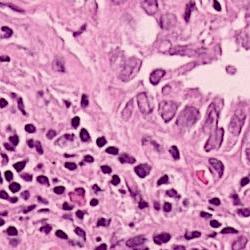} \\ 
		(a) image
	\end{minipage}
	\begin{minipage}{0.24\linewidth}
		\centering 	
		\includegraphics[width=\textwidth]{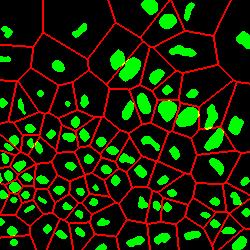} \\
		(b) true mask
	\end{minipage}
	\begin{minipage}{0.24\linewidth}
		\centering 
		\includegraphics[width=\textwidth]{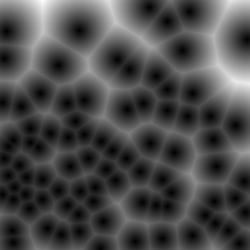} \\
		(c) dist map
	\end{minipage}
	\caption{Label generation. (a) image, (b) ground-truth nuclei masks (in green) and Voronoi edges (in red), (c) distance map.} \label{fig:s2}	
\end{figure}

A Voronoi diagram is a partitioning of a plane into convex polygons (Voronoi cells) according to the distance to a set of points in the plane. There is exactly one point (seed point) in each cell and all points in a cell are closer to its seed point than other seed points. In our task, the detected points in an image can be treated as seed points to calculate the Voronoi diagram as shown in Fig.~\ref{fig:overview}(b). For each Voronoi cell, assuming that the corresponding nucleus is located within the cell, then the Voronoi edges separate all nuclei well and the edge pixels belong to the background. This assumption holds for most of the nuclei because the detected points are around the centers and nuclear shapes are nearly convex (Fig.~\ref{fig:s2}(b)).

Assigning the Voronoi edges as background pixels and the detected points (dilated with a disk kernel of radius 2) as nuclei pixels, we obtain the Voronoi point-edge label (Fig.~\ref{fig:s2}). All other pixels are ignored during training. Note that although the pixels on the Voronoi edge between two touching nuclei may not necessarily be background, the edges are still helpful in guiding the network to separate the nuclei. The Voronoi labels aim to segment the central parts of nuclei and are not able to extract the full masks, because they lack the information of nuclear boundaries and shapes. To overcome this weakness, we generate another kind of labels that contain this complementary information.

\subsubsection{Cluster labels}
Considering the color difference between nuclei and background pixels, it is feasible to perform a rough segmentation using clustering methods. We choose $k$-means clustering to extract both nuclei and background pixels from the image, and generate the cluster labels. Given an image $\boldsymbol{x}$ with $N$ pixels ($x_1, x_2, \cdots, x_N$), $k$-means clustering aims to partition the $N$ pixels into $k$ clusters $\boldsymbol{S}=(S_1, S_2, \cdots, S_k)$ according to the feature vector $\boldsymbol{f}_{x_i}$ of each pixel $x_i$, such that the sum of within-cluster variances is minimized:
\begin{equation}
\arg\min_{\boldsymbol{S}}\sum_{i=1}^k\sum_{x\in S_i} \left\|\boldsymbol{f}_x - \boldsymbol{c}_i\right\|^2.
\end{equation}
We use $k$-means to divide all pixels into $k=3$ clusters: nuclei, background and ignored. The cluster that has maximum overlap with points label is considered as nuclei, and the cluster that has minimum overlap with the dilated points label is considered as background. The remaining one is the ignored class. The pixels of the ignored class are often located around the nuclear boundaries, which are hard for a clustering method to assign correct labels.

For the feature vector $\boldsymbol{f}$, color is a straightforward choice. However, clustering with color will result in wrong assignments for pixels inside some nuclei that have non-uniform colors. To cope with this issue, we add a distance feature. In a distance map (Fig.~\ref{fig:s2}(c)), each value indicates the distance of that pixel to the closest nuclear point and therefore incorporates the spatial information. In particular, the pixels that belong to nuclei should be close enough to points in the label while background pixels are relatively far from those points. The distance map can be calculated by the distance transform of the complement image of detected points. Combining the distance value $d_i$ with the RGB color values $(r_i, g_i, b_i)$ as the feature vector $\boldsymbol{f}_{x_i}=(\hat{d_i}, \hat{r_i}, \hat{g_i}, \hat{b_i})$ in $k$-means clustering, we obtain the initial cluster labels.  $\hat{d_i}$ is the clipped value by truncating large values to 20 and $\hat{r_i}$, $\hat{g_i}$, $\hat{b_i}$ are normalized values such that every feature has similar value range. 

The cluster label (Fig.~\ref{fig:overview}(b)) is generated by refining the clustering result with morphological dilation and erosion, which are done separately in each Voronoi cell to avoid connecting close nuclei. 
The cluster labels have more shape information about the nuclei compared with the Voronoi labels, at the expense of more errors and uncertainties. We argue that these two types of labels are complementary to each other and will jointly lead to better results.

\subsection{Training deep neural networks with pixel-level labels}
Once we have the two types of pixel-level labels, we are able to train a deep convolutional neural network for nuclei segmentation. The network structure is the same as that in nuclei detection. It outputs two probability maps of background and nuclei, which are used to calculate two cross entropy losses with respect to the cluster label $\mathcal{L}_{cluster}$ and Voronoi label $\mathcal{L}_{vor}$:
\begin{equation}
\mathcal{L}_{cluster/vor}(\boldsymbol{y}, \boldsymbol{t}) = - \frac{1}{|\Omega|} \sum_{i\in\Omega} \left[ t_i \log y_i + (1-t_i)\log(1-y_i)\right],
\end{equation}
where $\boldsymbol{y}$ is the probability map, $\boldsymbol{t}$ is the cluster label or Voronoi label, and $\Omega$ is the set consisting of non-ignored pixels. The final loss is 
\begin{equation}\label{eqn:loss_ce}
\mathcal{L}_{ce} = \alpha \mathcal{L}_{vor} + (1-\alpha)\mathcal{L}_{cluster},
\end{equation}
where $\alpha$ is a balancing parameter.

\subsection{Model refinement using dense CRF loss}
The model trained using the two types of labels is able to predict the mask of individual nucleus with high accuracy. To further improve the performance, we refine the nuclear boundaries with the dense CRF loss. Previously, post-processing such as region growing~\cite{kumar2017dataset}, graph search~\cite{yang2018boxnet} or dense CRF~\cite{chen2014semantic} is often utilized to refine the segmentation results. These algorithms introduce more computational complexity, making them unsuitable for the processing of high resolution Whole Slide Images. To solve this problem, similar to~\cite{tang2018regularized}, we embed the dense CRF into the loss function during training to improve the accuracy. The loss function is not calculated during inference, and therefore will not introduce additional computational cost after training.

Let $\tilde{\boldsymbol{y}}=(\tilde{y}_1, \tilde{y}_2, \cdots, \tilde{y}_N)$ denote the predicted label (0 for background and 1 for nuclei) from probability maps $\boldsymbol{y}$ and $\boldsymbol{t}$ be the label. The dense CRF is to minimize the energy function:
\begin{equation}
E(\tilde{\boldsymbol{y}}, \boldsymbol{t})=\sum_{i} \phi(\tilde{y}_i, t_i) + \sum_{i,j}\psi(\tilde{y}_i, \tilde{y}_j),
\end{equation}
where $\phi$ is the unary potential that measures how likely a pixel belongs to a certain class, and $\psi$ is the pairwise potential that measures how different a pixel's label is from all other pixels' in the image. The unary term is replaced with the cross entropy loss $\mathcal{L}_{ce}$. The pairwise potential usually has the form:
\begin{equation}
\psi(\tilde{y}_i, \tilde{y}_j) = \mu(\tilde{y}_i, \tilde{y}_j)W_{ij} =  \mu(\tilde{y}_i, \tilde{y}_j)\sum_{m=1}^K w_m k_m(\tilde{\boldsymbol{f}}_i, \tilde{\boldsymbol{f}}_j),
\end{equation}
where $\mu$ is a label compatibility function, $W_{ij}$ is the affinity between pixels $i,j$ and is often calculated by the sum of Gaussian kernels $k_m$. Here we choose $\mu$ as the Potts model, i.e., $\mu(\tilde{y}_i, \tilde{y}_j)=[\tilde{y}_i\ne\tilde{y}_j]$, and bilateral feature vector $\tilde{\boldsymbol{f}}_i=\left(\frac{p_i}{\sigma_{pq}}, \frac{q_i}{\sigma_{pq}}, \frac{r_i}{\sigma_{rgb}},\frac{g_i}{\sigma_{rgb}}, \frac{b_i}{\sigma_{rgb}}\right)$ that contains both location and color information. $\sigma_{pq}$ and $\sigma_{rgb}$ are Gaussian bandwidth. 

To adapt the energy function to a differentiable loss function, we relax the pairwise potential as~\cite{tang2018regularized}: 
\begin{equation}
\psi(\tilde{y}_i, \tilde{y}_j) = \tilde{y}_i (1-\tilde{y}_j)W_{ij}.
\end{equation}
Therefore, the dense CRF loss can be expressed as:
\begin{equation}
\mathcal{L}_{crf}(\boldsymbol{y}, \boldsymbol{t}_{cluster}, \boldsymbol{t}_{vor}) = \mathcal{L}_{ce}(\boldsymbol{y}, \boldsymbol{t}_{cluster}, \boldsymbol{t}_{vor}) + \beta \mathcal{L}_{pair}(\boldsymbol{y}),
\label{eqn:loss_crf}
\end{equation}
where $\mathcal{L}_{pair}(\boldsymbol{y})=\sum_{i,j}y_i (1-y_j)W_{ij}$ is the pairwise potential loss and $\beta$ is the weighting factor. The CRF loss is used to fine-tune the trained model. Due to the large number of pixels in an image, the cost of directly computing the affinity matrix $W=[W_{ij}]$ is prohibitive. For instance, there are $N^2=1.6\times10^9$ elements in $W$ for an image of size $200\times200$ that has $N=40000$ pixels. We adopt fast mean-field inference based on high-dimensional filtering~\cite{adams2010fast} to compute the pairwise potential term.

\section{Experiments}\label{sec:exp}

To validate the proposed framework, we conduct experiments on two datasets of H\&E stained histopathology images.

\subsection{Datasets}
\subsubsection{Lung Cancer (LC) dataset}
We generated this dataset by extracting 40 images of size $900\times900$ from 8 lung adenocarcinoma or lung squamous cell carcinoma cases, i.e., H\&E stained WSIs with 20x magnification. They are split into the training, validation and test sets, consisting of 24, 8 and 8 images, respectively. 24401 nuclei are annotated with masks. 

\subsubsection{Multi-Organ (MO) dataset}
It is a public dataset released by Kumar et al.~\cite{kumar2017dataset}, and consists of 30 images of size $1000\times1000$ which are taken from multiple hospitals including a diversity of nuclear appearances from seven organs~\cite{kumar2017dataset}. The variability in this dataset is large because of the heterogeneity between organs and cancer types. There are 12, 4 and 14 images in training, validation and test sets.

Both datasets have full mask annotation. We use the bounding box centers of the nuclear masks as ground-truth for the detection. To generate the partial points annotation in the training set, we randomly sample a certain ratio of points.

\subsection{Detection using partial points annotation}
The aim of this experiment is to detect all nuclei in an image using the model trained with partial points annotation. 

\subsubsection{Evaluation metrics}
We adopt the common metrics for detection tasks: precision (P), recall (R) and F1 score. They are defined as: $P=TP/(TP+FP)$, $R=TP/(TP+FN)$, $F1=2TP/(2TP+FP+FN)$, where $TP, FP, FN$ are the number of true positives, false positives and false negatives, respectively. A detected nucleus is a true positive if it locates in a circle centered at a nuclear centroid with $r$-pixel radius, otherwise it is a false positive. The ground-truth points which have no corresponding detection are false negatives. If there are multiple detected points for the same ground-truth point, only the closest one is considered as a true positive. $r$ is the rough average nuclear radius computed using the validation set. We set $r=8$ for the LC dataset and $r=11$ for the MO dataset. We also adopt the mean ($\mu_d$) and standard deviation ($\sigma_d$) of the detection distance error to evaluate the localization accuracy. They are defined as 
\begin{equation}
\mu_d = \frac{1}{N_{TP}}\sum_{i=1}^{N_{TP}}d_i, \quad \sigma_d = \sqrt{\frac{1}{N_{TP}}\sum_{i=1}^{N_{TP}}\left(d_i-\mu_d\right)^2}
\end{equation}
where $N_{TP}$ is the total number of true positive detected nuclei in all test images, $d_i$ is the Euclidean distance between the $i$-th groundtruth point and the true positive detection.

\begin{table}[t]
	\caption{Nuclei detection results of different strategies on LC and MO datasets using 10\% partial points annotation.}\label{tab:detect:strategy}
	\centering
	\begin{tabular}{llccccc}
		\toprule
		Dataset & Method & P & R & F1 & $\mu_d$ & $\sigma_d$  \\ \midrule
		\multirow{5}{*}{LC}   & Full & \underline{0.8767} & \underline{0.9141} & \underline{0.8950} & \underline{1.14} & \underline{1.03} \\
		&  GM		& 0.6322	& 0.6337 	& 0.6329 	& 2.34 	& 2.00 \\
		& ext-GM	& 0.7483	& \textbf{0.9306} & 0.8296 & 1.74 & 1.45 \\
		& ST-nu   	& 0.7505	& 0.9016	& 0.8192 & 1.70 & 1.49 \\
		& ST-bg  & \textbf{0.8605} & 0.9171 & \textbf{0.8879}  & \textbf{1.42}  & \textbf{1.22} \\ \midrule
		\multirow{5}{*}{MO}   & Full & \underline{0.8420} & \underline{0.8665} & \underline{0.8541} & \underline{2.68} & \underline{2.00} \\
		& GM	&  0.5574	& 0.7650 	& 0.6449	& 3.85	& 2.68 \\
		& ext-GM & 0.7932	& \textbf{0.8471} 	 & 0.8193	 & \textbf{2.87}	 & 2.18 \\
		& ST-nu	& 0.7754	 &  0.8138	& 0.7941 	 & 2.95	 & 2.17 \\
		& ST-bg	& \textbf{0.8238}	& 0.8328	 & \textbf{0.8282}	& 2.90 & \textbf{2.07} \\
		\bottomrule
	\end{tabular}
\end{table}

\begin{table}[t]
	\caption{Nuclei detection results on LC and MO datasets using different ratios of annotation.}\label{tab:detect:ratio}
	\centering
	\begin{tabular}{llccccc}
		\toprule
		Dataset & Method & P & R & F1 & $\mu_d$ & $\sigma_d$  \\ \midrule
		\multirow{5}{*}{LC}   & Full & \underline{0.8767} & \underline{0.9141} & \underline{0.8950} & \underline{1.14} & \underline{1.03} \\
		&  5\%  & 0.8564	&  0.9171 	& 0.8857 	& 1.53	& 1.30 \\
		& 10\%	& \textbf{0.8605}	&  0.9171 & 0.8879  & 1.42 & 1.22 \\
		& 25\%   	&  0.8517	&  0.9399	& \textbf{0.8936}  & 1.35 & 1.17 \\
		& 50\%  &  0.8502 & \textbf{0.9414} & 0.8935 & \textbf{1.30}  &  \textbf{1.12} \\ \midrule
		\multirow{5}{*}{MO}   & Full & \underline{0.8420} & \underline{0.8665} & \underline{0.8541} & \underline{2.68} & \underline{2.00} \\
		& 5\%	&   0.8021	&  0.8441 	&  0.8226	&  3.04	&  2.13 \\
		& 10\% & 0.8238 	&  0.8328 	 & 0.8282	 & 2.90	 & 2.07 \\
		& 25\%	& \textbf{0.8259}	 &  0.8440	&  0.8349 	 & 2.97	 & 2.05 \\
		& 50\%	& 0.8237	& \textbf{0.8821}	 &  \textbf{0.8519}	& \textbf{2.76} & \textbf{2.03} \\
		\bottomrule
	\end{tabular}
\end{table}

\subsubsection{Implementation details}
Color normalization~\cite{reinhard2001color} is applied to all images to remove color variations caused by staining. We extract 16 image patches of size 250$\times$250 from each training image, and randomly crop 224$\times$224 patches as network inputs. Other data augmentations are conducted including random crop, scale, rotation, flipping, and affine transformations. The encoder part of the network is initialized with the pre-trained parameters. The model is trained using the Adam optimizer~\cite{kingma2014adam} for 80 epochs in initial training and each round of self-training. The batch size is 16. The learning rate is 1e-4 for LC dataset and 1e-3 for MO dataset. The parameters in the extended Gaussian mask in Eqn.~(\ref{eqn:1}) are $r_1=8, r_2=16, \sigma=2$ for LC dataset and $r_1=11, r_2=22, \sigma=2.75$ for MO dataset. During inference, models from round 3 of self-training are used to evaluate on the test data, since it is already converged.

\subsubsection{Results and discussion}

We compare different strategies mentioned in Section~\ref{sec:detection}:
\begin{itemize}
	\item Full: fully-supervised training using all annotated nuclei.
	\item GM: initial training using simple Gaussian masks from partial points annotation, i.e., no ignored pixels. 
	\item ext-GM: initial training using our proposed extended Gaussian masks from partial points annotation.
	\item ST-nu: updating the label by adding detected nuclei in the self-training step of our method.
	\item ST-bg: updating the label by propagating background pixels in the self-training step of our method.
\end{itemize}
For both ST-nu and ST-bg, ext-GM is used in the initial training. The detection results using 10\% points in each training image are reported in Table~\ref{tab:detect:strategy}.

\paragraph{Initial training strategies}
In the two initial training strategies, the ext-GM achieves better performance on both datasets. For GM, all the unlabeled nuclei are treated as background, which biases the training and guides the network to predict pixels as background more aggressively. Our extended Gaussian masks force the model to focus on the areas around the labeled points. As a result, the trained model is able to make correct predictions in similar unlabeled regions.

\paragraph{Self-training strategies}
In the self-training stage, compared with the results of the first stage (ext-GM), updating the nuclei (ST-nu) decreases the performance. The reason is that the number of false positives in the newly added nuclei is comparable to that of the labeled nuclei, resulting in a negative effect on training. In contrast, our background propagation strategy (ST-bg) keeps the labeled nuclei unchanged and gradually increases the background area (as shown in Fig.~\ref{fig:d2}), which doesn't introduce false positives during training. Therefore, it can improve both recall and precision, resulting in a much higher F1 score and localization accuracy.

\paragraph{Comparison to fully-supervised case}
Compared with the results of full annotation (Full), our method (ST-bg) can achieve comparable performance while using much fewer annotation data. On the LC dataset, the precision, recall and F1 are 98.2\%, 100.3\%, 99.2\% of the fully-supervised results, respectively. And these numbers are 97.8\%, 96.1\%, 97.0\% respectively on the MO dataset.
Besides, the localization error $\mu_d \pm \sigma_d$ is also very close to that using full annotation on both datasets.

The typical qualitative results of different training strategies from both datasets using 10\% points are shown in Fig.~4.

\begin{figure*}[t]
	\centering
	\includegraphics[width=0.162\textwidth]{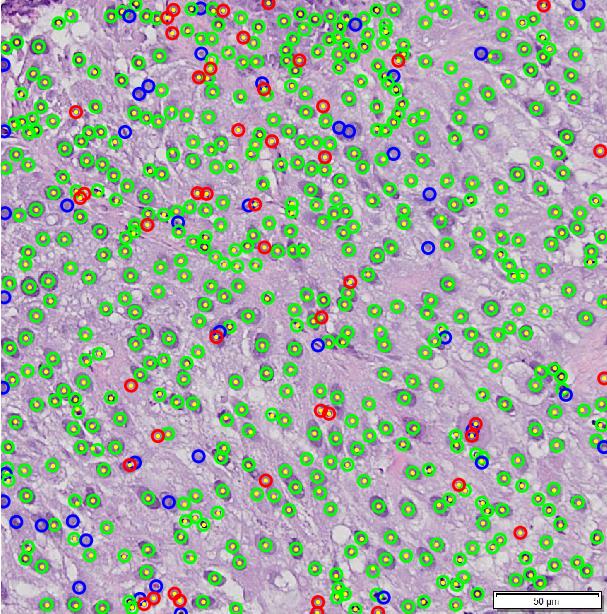} \hspace{-0.07in}
	\includegraphics[width=0.162\textwidth]{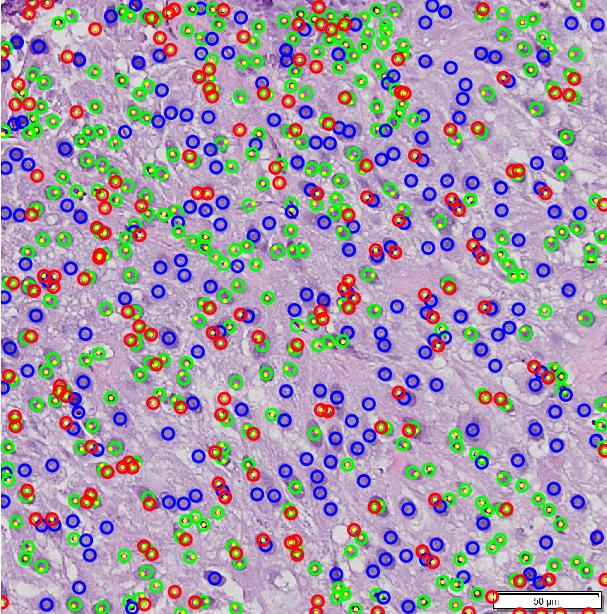} \hspace{-0.07in}
	\includegraphics[width=0.162\textwidth]{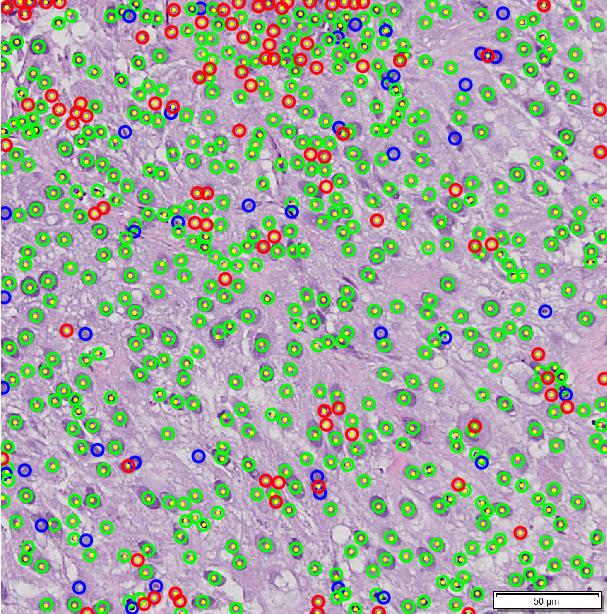} \hspace{-0.07in}
	\includegraphics[width=0.162\textwidth]{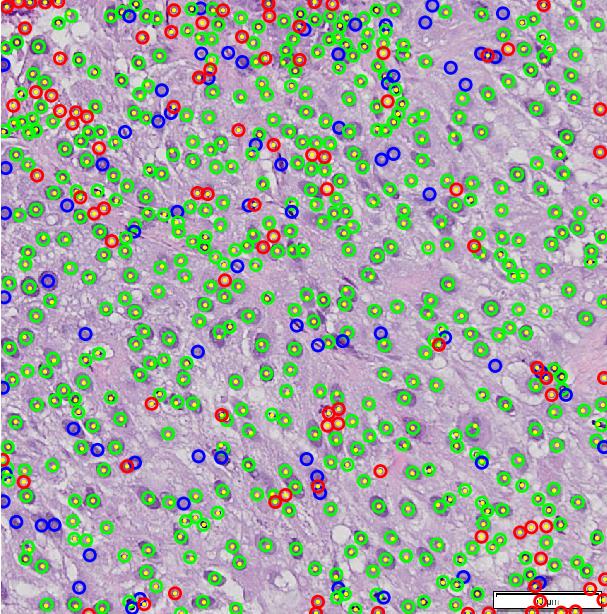} \hspace{-0.07in}
	\includegraphics[width=0.162\textwidth]{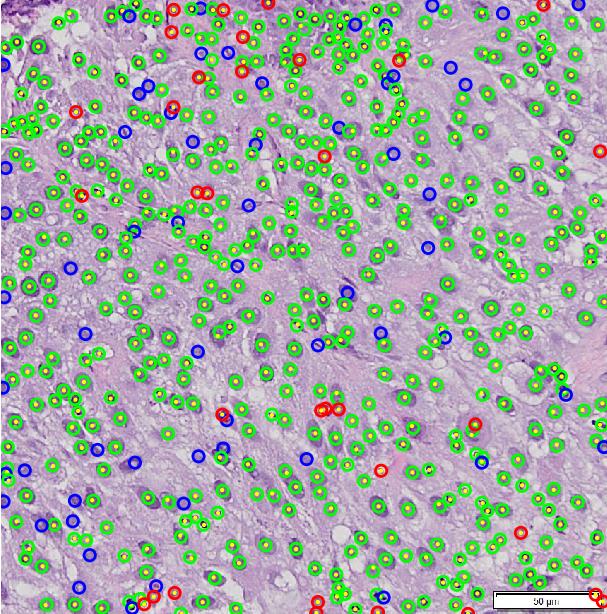} \\ \vspace{0.02in}
	\includegraphics[width=0.162\textwidth]{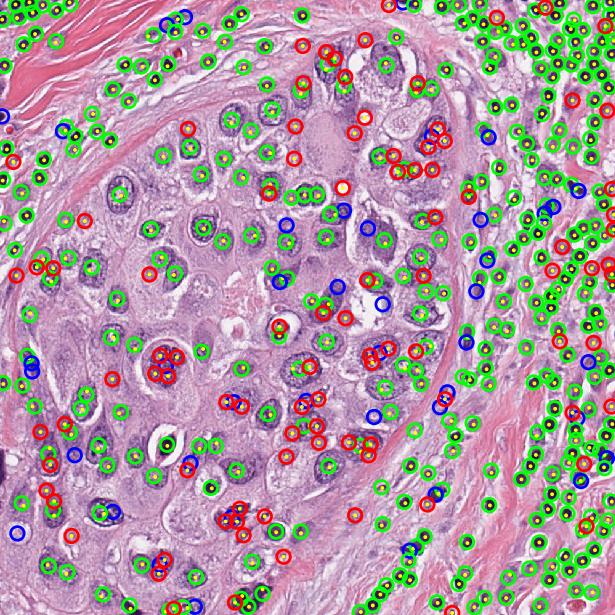} \hspace{-0.07in}
	\includegraphics[width=0.162\textwidth]{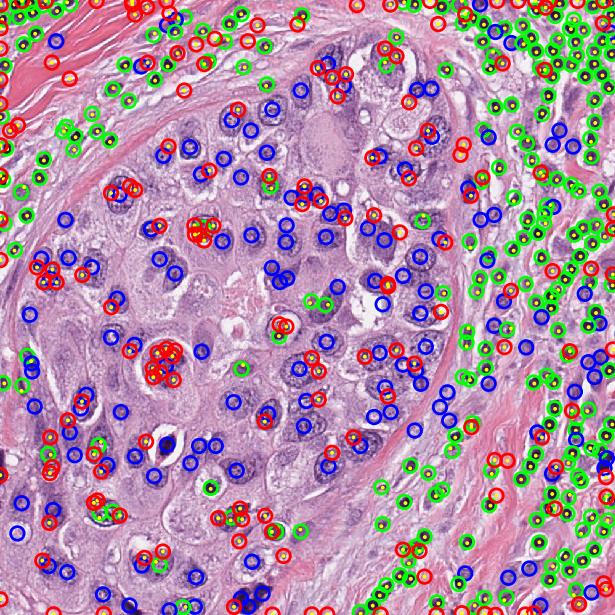} \hspace{-0.07in}
	\includegraphics[width=0.162\textwidth]{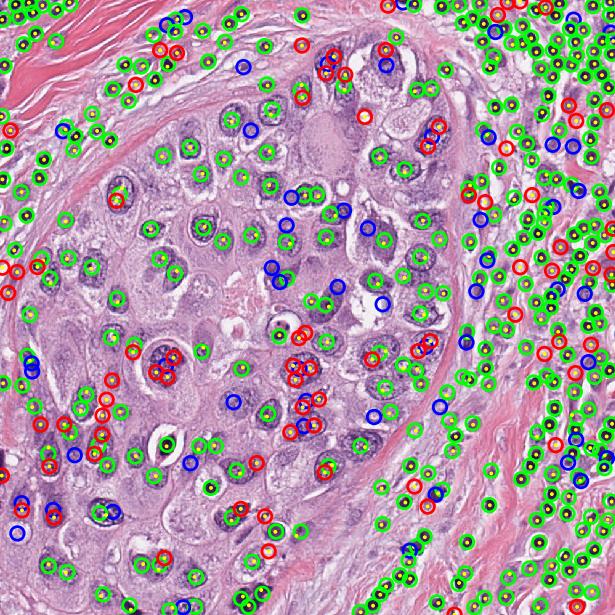} \hspace{-0.07in}
	\includegraphics[width=0.162\textwidth]{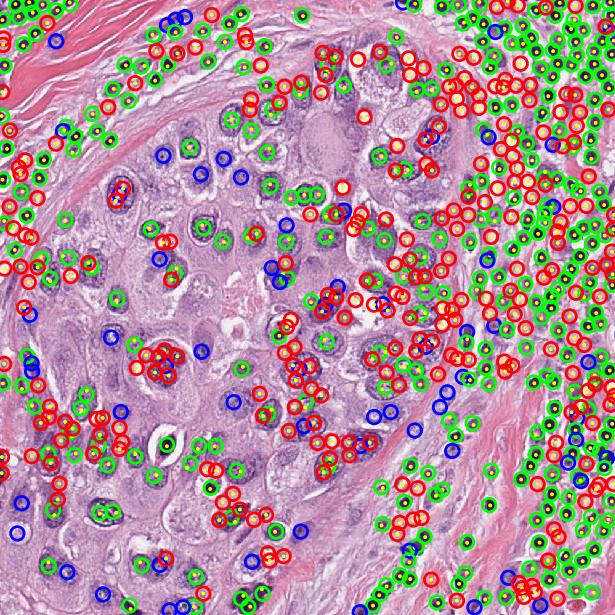} \hspace{-0.07in}
	\includegraphics[width=0.162\textwidth]{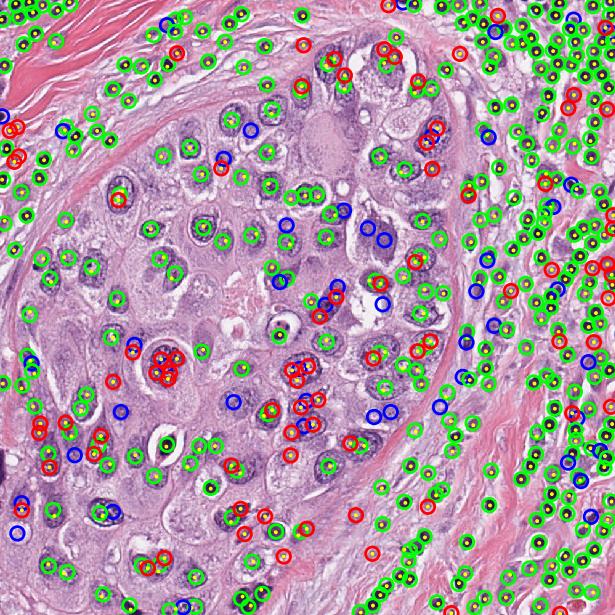} \\
	\begin{minipage}{0.162\textwidth}	\centering(a) Full	    \end{minipage}
	\begin{minipage}{0.16\textwidth}	\centering(b) GM    	\end{minipage} 
	\begin{minipage}{0.162\textwidth}	\centering(c) ext-GM    \end{minipage} 
	\begin{minipage}{0.162\textwidth} 	\centering(d) ST-nu 	\end{minipage} 
	\begin{minipage}{0.162\textwidth}	\centering(e) ST-bg 	\end{minipage} 
	\caption{Typical detection results of different strategies from Lung Cancer dataset (first row) and Multi-Organ dataset (second row) using 10\% points. (a) Fully supervised training, (b) initial training using simple Gaussian mask, (c) initial training using extended Gaussian mask, (d) self-training using nuclei prediction, (e) self training using background propagation. Yellow dots are the detected nuclei. Green, blue and red circles represent ground-truth with correct detection (TP), ground-truth without correct detection (FN) and false positive detection (FP), respectively.}
	\label{fig:det-results}
\end{figure*}

\subsubsection{Different ratios of annotation}
To explore how the proposed nuclei detection method behaves when the ratio of points changes, we trained with 5\%, 10\%, 25\% and 50\% partial points annotation. The results are shown in Table~\ref{tab:detect:ratio}. The detection accuracy (F1-score) increases and localization error ($\mu_d\pm\sigma_d$) decreases as more annotations are available. With 50\% points annotated, the performance is nearly the same as that using full annotation. Even with only 5\% annotation, the F1 score of our method can reach 99.0\% of that using full annotation on LC dataset and 96.3\% on the MO dataset, which substantiates the effectiveness of our algorithm. The results of MO dataset are slightly worse compared with those on the LC dataset because the MO dataset is more challenging due to the diversity in nuclear size and appearance.

\subsection{Segmentation using ground-truth points}
In this subsection, we present the experimental results of our segmentation method based on all ground-truth points. We first try to obtain the optimal value of $\alpha$ in Eqn.~(\ref{eqn:loss_ce}) and discuss the effects of two types of labels, then show the effects of the dense CRF loss, and finally compare our results with fully supervised ones.

\subsubsection{Evaluation metrics}
Four metrics are used to evaluate the segmentation performance. At pixel-level, we use pixel accuracy and pixel-level F1 score.
Because nuclei segmentation is an instance segmentation task, two object-level metrics are also used: object-level Dice coefficient~\cite{sirinukunwattana2015stochastic} (Dice$_{obj}$) and the Aggregated Jaccard Index (AJI)~\cite{kumar2017dataset}. Dice$_{obj}$ is defined as
\begin{equation}
\begin{aligned}
Dice_{obj}(\mathcal{G}, \mathcal{S}) &= \frac{1}{2}\sum_{i=1}^{n_{\mathcal{G}}}\gamma_i Dice(G_i, S^*(G_i))\\
&+\frac{1}{2}\sum_{j=1}^{n_{\mathcal{S}}}\eta_j Dice(G^*(S_j), S_j)
\end{aligned}
\end{equation}
where $\gamma_i$, $\eta_j$ are the weights related to object areas, $\mathcal{G}$, $\mathcal{S}$ are the set of ground-truth objects and segmented objects, $S^*(G_i)$, $ G^*(S_i)$ are the segmented object that has maximum overlapping area with $G_i$ and ground-truth object that has maximum overlapping area with $S_i$, respectively. The correspondence is established if the overlapping area of two objects are more than 50\%. 
The AJI is defined as
\begin{equation}
AJI = \frac{\sum_{i=1}^{n_\mathcal{G}} |G_i \cap S(G_i)|}{\sum_{i=1}^{n_\mathcal{G}}|G_i \cup S(G_i)| + \sum_{k\in K}|S_k|}
\end{equation}
where $S(G_i)$ is the segmented object that has maximum overlap with $G_i$ based on the Jaccard index, $K$ is the set containing segmentation objects that have not been assigned to any ground-truth object.

\subsubsection{Implementation details}
In weakly supervised settings we train a model for 100 epochs with a learning rate of 1e-4, and fine-tune the model using the dense CRF loss for 20 epochs with a learning rate of 1e-5. 
In fully supervised settings, we train 200 epochs using binary masks with a learning rate of 1e-4. The validation set is used to select the best model for testing.

\subsubsection{Results and discussion}

\begin{figure*}[t]
	\centering
	\includegraphics[width=0.14\textwidth]{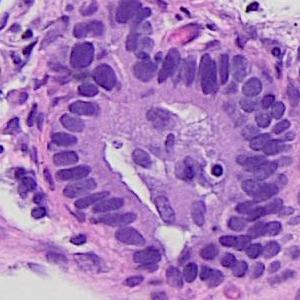} \hspace{-0.08in}
	\includegraphics[width=0.14\textwidth]{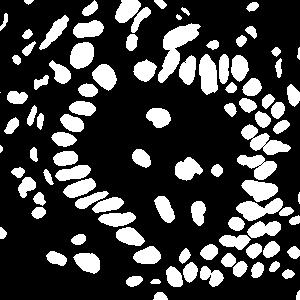}  \hspace{-0.08in}
	\includegraphics[width=0.14\textwidth]{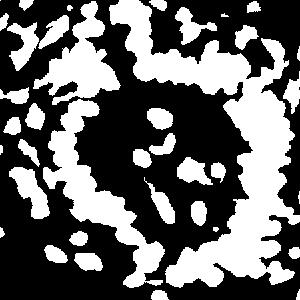}  \hspace{-0.08in}
	\includegraphics[width=0.14\textwidth]{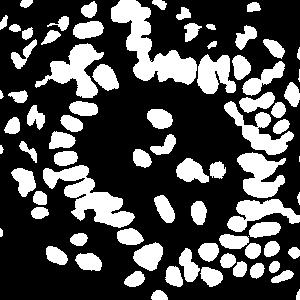}  \hspace{-0.08in}
	\includegraphics[width=0.14\textwidth]{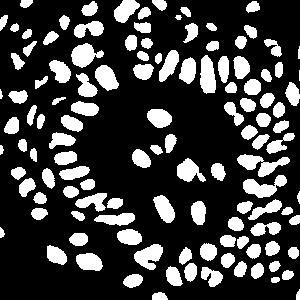}  \hspace{-0.08in}
	\includegraphics[width=0.14\textwidth]{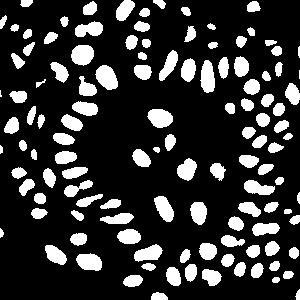}  \hspace{-0.08in}
	\includegraphics[width=0.14\textwidth]{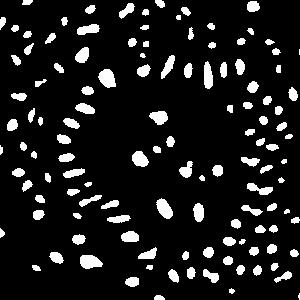}  \\ \vspace{0.02in}
	\includegraphics[width=0.14\textwidth]{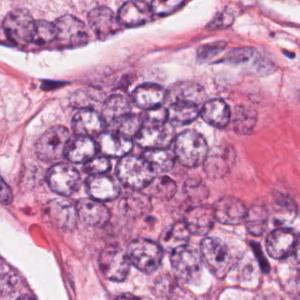} \hspace{-0.08in}
	\includegraphics[width=0.14\textwidth]{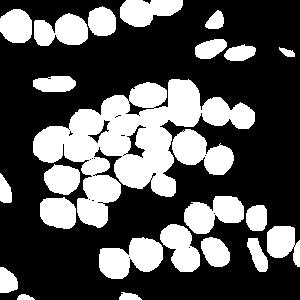}  \hspace{-0.08in}
	\includegraphics[width=0.14\textwidth]{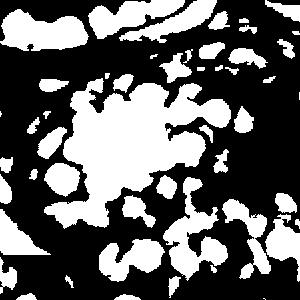}  \hspace{-0.08in}
	\includegraphics[width=0.14\textwidth]{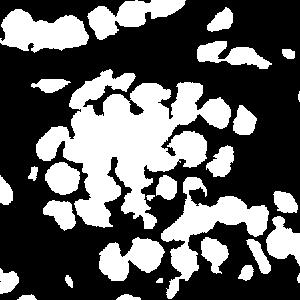}  \hspace{-0.08in}
	\includegraphics[width=0.14\textwidth]{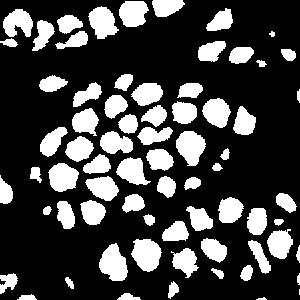}  \hspace{-0.08in}
	\includegraphics[width=0.14\textwidth]{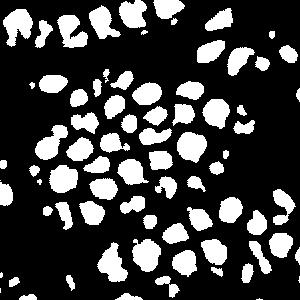}  \hspace{-0.08in}
	\includegraphics[width=0.14\textwidth]{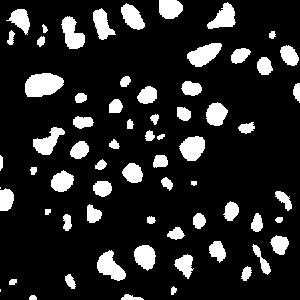}  \\
	\begin{minipage}[t]{0.135\textwidth}\centering	(a) \footnotesize{image}	\end{minipage}\hspace{-0.01in}
	\begin{minipage}[t]{0.135\textwidth}\centering	(b) \footnotesize{true mask} \end{minipage}\hspace{-0.01in}
	\begin{minipage}[t]{0.135\textwidth}\centering	(c) \footnotesize{$\alpha=0$} \\ \footnotesize{Cluster only}\end{minipage}\hspace{-0.01in}
	\begin{minipage}[t]{0.135\textwidth}\centering	(d) \footnotesize{$\alpha=0.25$} \end{minipage}\hspace{-0.01in}
	\begin{minipage}[t]{0.135\textwidth}\centering	(e) \footnotesize{$\alpha=0.50$} \end{minipage}\hspace{-0.02in}
	\begin{minipage}[t]{0.135\textwidth}\centering	(f) \footnotesize{$\alpha=0.75$} \end{minipage}\hspace{-0.02in}
	\begin{minipage}[t]{0.135\textwidth}\centering	(g) \footnotesize{$\alpha=1$} \\ \footnotesize{Voronoi only} \end{minipage}\hspace{-0.01in}
	\caption{Typical results using different weights ($\alpha$) of the cluster and Voronoi labels on LC dataset (top row) and MO dataset (bottom row). (a) image, (b) ground-truth full mask, (c)-(g) are results using different $\alpha$ values. $\alpha=0$ means using only the cluster label and $\alpha=1$ means using only the Voronoi label.} \label{fig:seg:diff-alpha}
\end{figure*}

\begin{figure}
	\centering
	\includegraphics[width=0.46\linewidth]{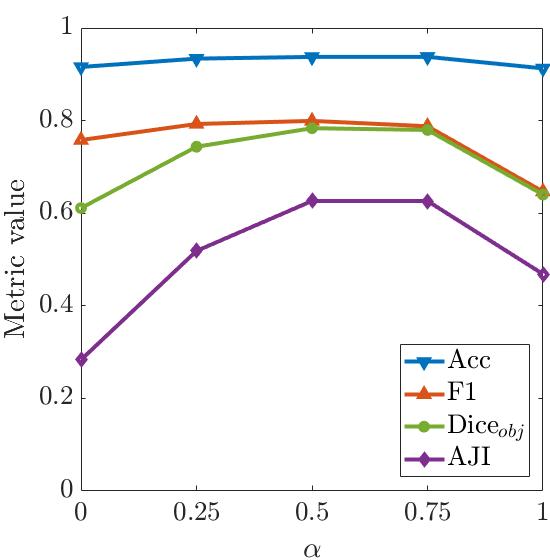}
	\includegraphics[width=0.46\linewidth]{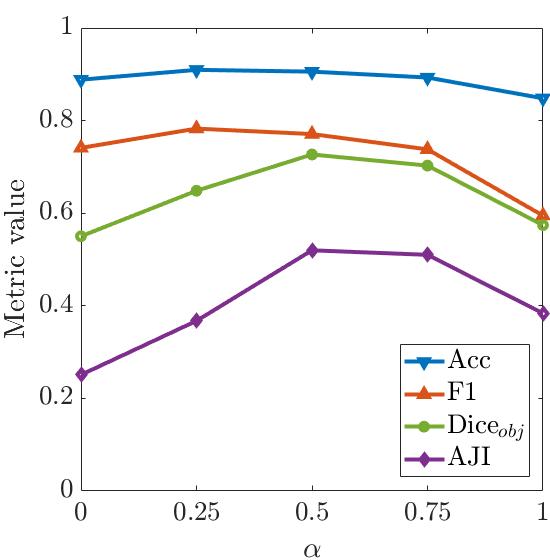}\\
	\begin{minipage}[t]{0.46\linewidth}\centering	(a) LC dataset	\end{minipage}
	\begin{minipage}[t]{0.46\linewidth}\centering	(b) MO dataset \end{minipage}
	\caption{The results of using different $\alpha$ values on LC and MO datasets.}
	\label{fig:seg:diff-alpha-metrics}
\end{figure}

\paragraph{The effects of two types of labels}
In order to explore how the two types of generated labels work on the model training, we change the values of $\alpha$ in Eqn.~(\ref{eqn:loss_ce}).
As $\alpha$ changes from 0 to 1, all four metrics increase in the beginning and then decrease (shown in Fig.~\ref{fig:seg:diff-alpha-metrics}). Compared to the results using only the cluster labels ($\alpha=0$), those with Voronoi labels ($\alpha=1$) are better in the object-level metrics, but worse in pixel-level metrics. This is because the model trained with Voronoi labels predicts the central parts of nuclei, resulting in small separated instances (Fig.~\ref{fig:seg:diff-alpha}(g)). While lacking the Voronoi edge information, the model using cluster labels is not able to separate close nuclei (Fig.~\ref{fig:seg:diff-alpha}(c)). In contrast, segmentation results using both labels (Fig.~\ref{fig:seg:diff-alpha}(d)-(f)) are better than those with either label alone, because they have both the shape information from the cluster label and the nuclei/background information from the Voronoi label. The best performance is achieved when $\alpha$ is around 0.5, thus we set $\alpha=0.5$ for all subsequent experiments.

\paragraph{The effects of dense CRF loss}
In the dense CRF loss, the Gaussian bandwidth parameters $\sigma_{pq}$ and $\sigma_{rgb}$ control the affinity between pixel pairs, thus having an impact on the effect of the loss along with the weight $\beta$. We perform an ablation study on the three parameters to show how their values affect the segmentation performance. 
The ranges are $\sigma_{pq}\in\{3,6,9,12,15\}$, $\sigma_{rgb}\in\{0.05, 0.1, 0.2, 0.3\}$ and $\beta\in\{0.05,0.01,0.005,0.001,0.0005,0.0001\}$. Among different value combinations, the best for LC dataset is $\sigma_{pq}=9$, $\sigma_{rgb}=0.2$, $\beta=0.001$, and that of MO dataset is $\sigma_{pq}=9$, $\sigma_{rgb}=0.1$, $\beta=0.005$.
\begin{figure}
	\centering
	\includegraphics[width=0.49\linewidth]{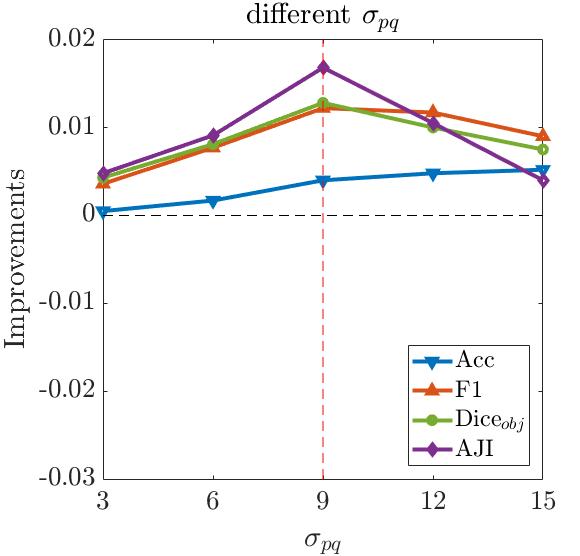}
	\includegraphics[width=0.49\linewidth]{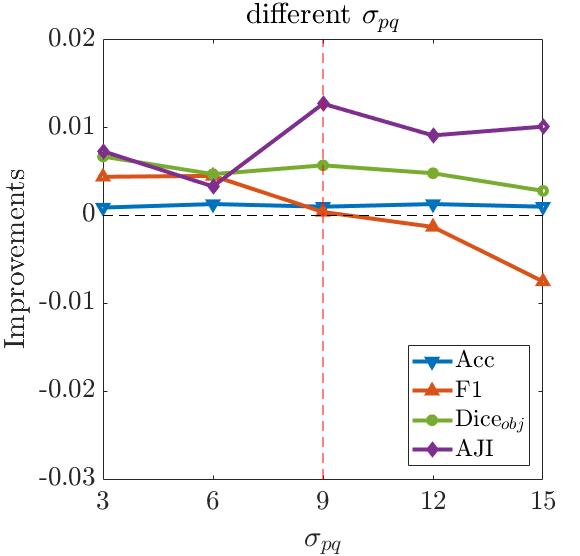}\\ \vspace{0.05in}
	\includegraphics[width=0.49\linewidth]{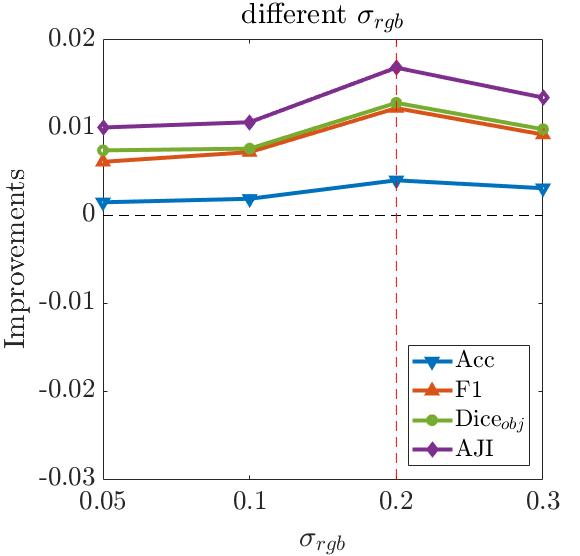}
	\includegraphics[width=0.49\linewidth]{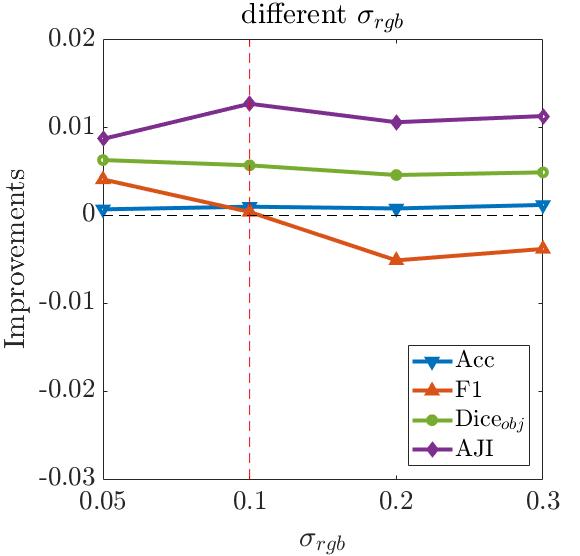}\\ \vspace{0.05in}
	\includegraphics[width=0.49\linewidth]{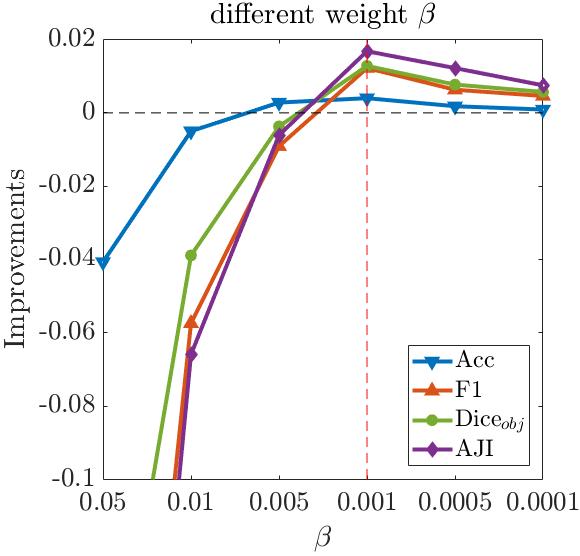}
	\includegraphics[width=0.49\linewidth]{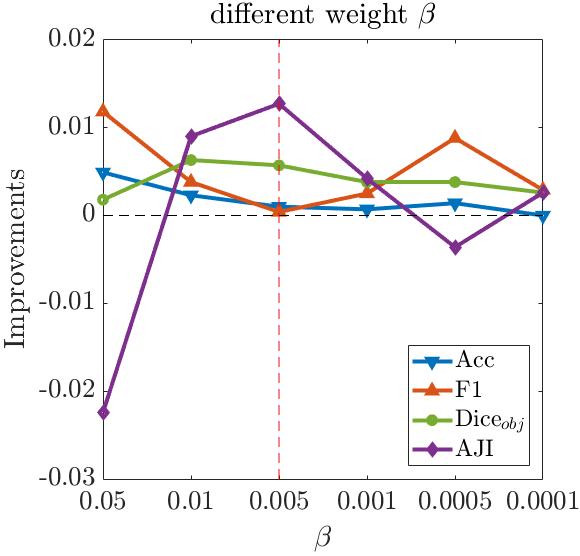}
	\begin{minipage}[t]{0.49\linewidth}\centering	(a) LC dataset	\end{minipage}
	\begin{minipage}[t]{0.49\linewidth}\centering	(b) MO dataset \end{minipage}
	\caption{The improvements over the baseline using different $\sigma_{pq}$, $\sigma_{rgb}$ and $\beta$ values in the CRF loss for LC and MO datasets. In each subfigure, the horizontal black dash line indicates the baseline's performance, and the vertical red dash line indicates the parameter value of the best combination.}
	\label{fig:seg:crf-diff-settings}
\end{figure}

The results of varying the value of one parameter based on the best combination are shown in Fig.~\ref{fig:seg:crf-diff-settings}. For simplicity, we plot the improvements of the finetuned models over the baseline.
When $\sigma_{pq}$ is small, only pixel pairs that are close enough have a large affinity. As a result, the current pixel's prediction is affected by local pixels, which leads to good results for small objects. On the contrary, a large $\sigma_{pq}$ is good for large objects. Taking all nuclei into account, $\sigma_{pq}$ should not be too small or too large, which is revealed by the results in Fig.~\ref{fig:seg:crf-diff-settings}. The rule is the same for $\sigma_{rgb}$. The weight $\beta$ adjusts the importance of the unary and pairwise potentials in the loss (Eqn.~(\ref{eqn:loss_crf})). A large $\beta$ emphasizes the pairwise relationship obtained from the image during the fine-tuning process, biasing the baseline model trained on the labels. Therefore, the performance degrades a lot, especially on the LC dataset. While a small $\beta$ makes the fine-tuning less effective.

\begin{table}[t]
	\centering
	\caption{Nuclei segmentation results on LC and MO datasets for our partial points annotation, fully-supervised and state-of-the-arts.}
	\label{tab:seg:ratio}
	\begin{tabular}{llcccc}
		\toprule		
		\multirow{2}{*}{Dataset} & \multirow{2}{*}{Method}  & \multicolumn{2}{c}{Pixel-level} & \multicolumn{2}{c}{Object-level}  \\	\cmidrule{3-6}
		& & Acc & F1 & Dice$_{obj}$ & AJI \\ \midrule
		\multirow{6}{*}{LC} & Fully-sup & \underline{0.9615} & \underline{0.8771} & \underline{0.8521} & \underline{0.6979} \\
		& GT points & 0.9427 & 0.8143 & 0.8021  & 0.6497\\ \cmidrule{2-6}
		& 5\%   & 0.9262 & 0.7612 & 0.7470 & 0.5742 \\
		& 10\%  & 0.9312 & 0.7700 & 0.7574 & 0.5754 \\
		& 25\%  & 0.9331 & 0.7768 & 0.7653 & 0.6003 \\
		& 50\%  & 0.9332 & 0.7819 & 0.7704 & 0.6120 \\ \midrule
		\multirow{8}{*}{MO} & CNN3~\cite{kumar2017dataset} & - & - & - & 0.5083\\
		& DIST~\cite{naylor2017nuclei} & - & 0.7623 & - & 0.5598 \\
		& Fully-sup & \underline{0.9194} & \underline{0.8100} & \underline{0.6763} & \underline{0.3919} \\ 
		& GT points & 0.9097 & 0.7716 & 0.7242 & 0.5174\\ \cmidrule{2-6}
		& 5\%   & 0.8951 & 0.7540 & 0.7015 & 0.4941 \\
		& 10\%  & 0.8997 & 0.7490 & 0.7033 & 0.5031 \\
		& 25\%  & 0.8966 & 0.7511 & 0.7087 & 0.5120 \\
		& 50\%  & 0.8999 & 0.7566 & 0.7157 & 0.5160 \\
		\bottomrule
	\end{tabular}
\end{table}

\paragraph{Results using ground-truth points}
With the above parameter settings, the results using all ground-truth points are shown in Table~\ref{tab:seg:ratio}. The segmentation performance of our weakly supervised method using all ground-truth points is close to that of the fully supervised models with the same network structure. On the Lung Cancer dataset, the gaps for accuracy, F1 score, Dice and AJI are 2.0\%, 7.2\%, 5.9\%, 6.9\%, respectively. On the MultiOrgan dataset, the gaps for accuracy and F1 score are 1.1\% and 4.7\%, respectively. However, the fully supervised model has very low Dice and AJI, since for fair comparison we didn't perform post-processing to separate the touching nuclei for any of the methods. The weakly supervised model is able to separate most of them due to the Voronoi labels while the fully supervised model failed to achieve this. Compared to the CNN3 method in~\cite{kumar2017dataset}, our method achieved a similar accuracy in terms of the AJI value. Compared to the state-of-the-art DIST method~\cite{naylor2018segmentation}, our approach has a higher pixel-level F1 score, but still has room for improvement on the nuclear shapes, as indicated by the AJI values.

\begin{figure*}[t]
	\centering
	\includegraphics[width=0.14\textwidth]{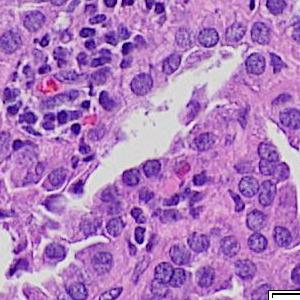} \hspace{-0.08in}
	\includegraphics[width=0.14\textwidth]{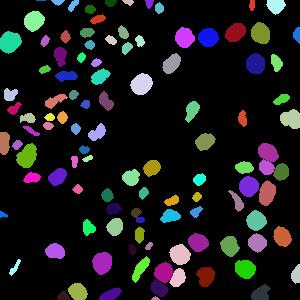}  \hspace{-0.08in}
	\includegraphics[width=0.14\textwidth]{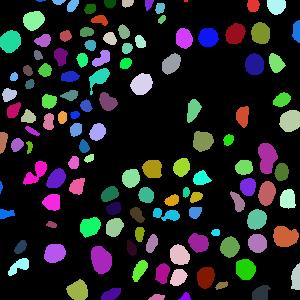}  \hspace{-0.08in}
	\includegraphics[width=0.14\textwidth]{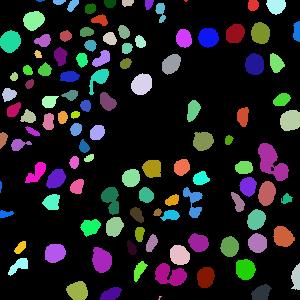}  \hspace{-0.08in}
	\includegraphics[width=0.14\textwidth]{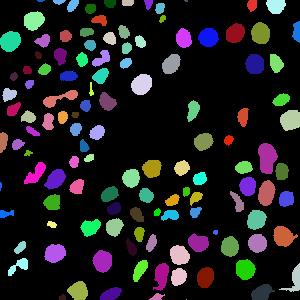}  \hspace{-0.08in}
	\includegraphics[width=0.14\textwidth]{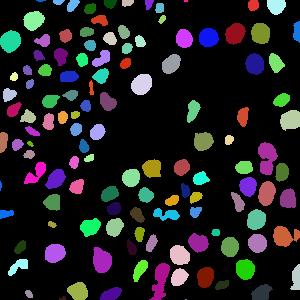}  \hspace{-0.08in}
	\includegraphics[width=0.14\textwidth]{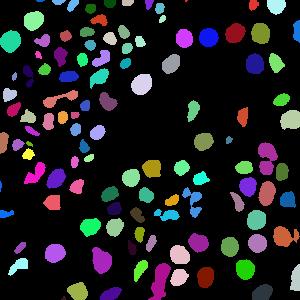} \\ \vspace{0.02in}
	\includegraphics[width=0.14\textwidth]{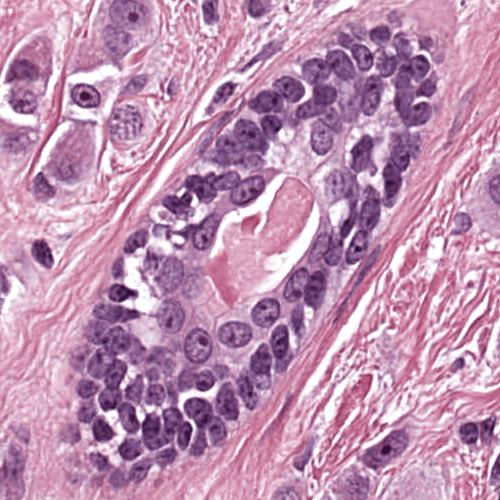} \hspace{-0.08in}
	\includegraphics[width=0.14\textwidth]{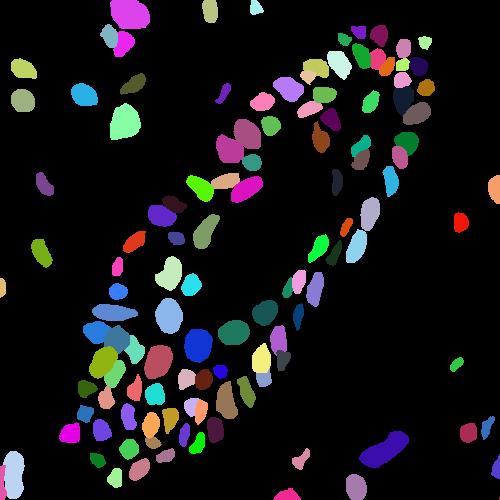}  \hspace{-0.08in}
	\includegraphics[width=0.14\textwidth]{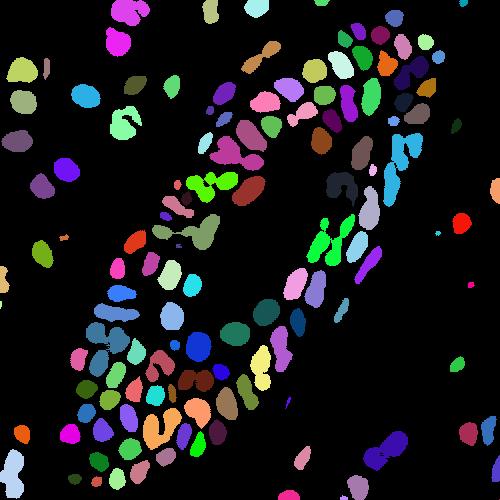}  \hspace{-0.08in}
	\includegraphics[width=0.14\textwidth]{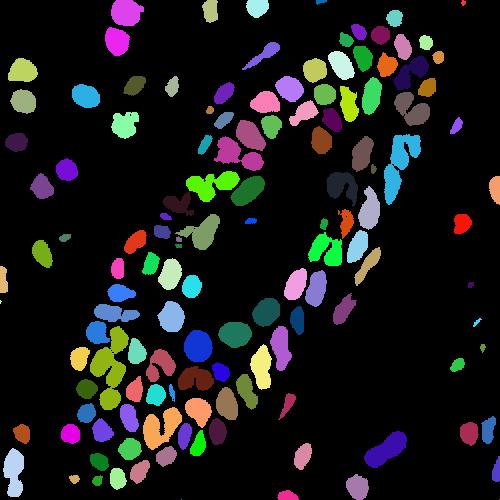}  \hspace{-0.08in}
	\includegraphics[width=0.14\textwidth]{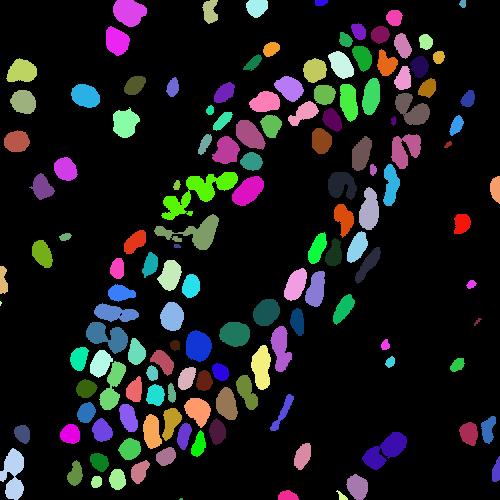}  \hspace{-0.08in}
	\includegraphics[width=0.14\textwidth]{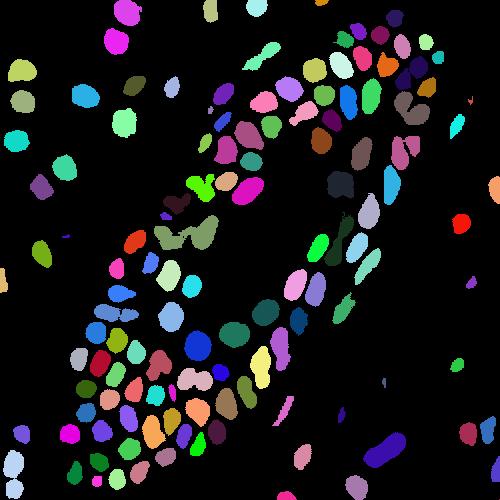}  \hspace{-0.08in}
	\includegraphics[width=0.14\textwidth]{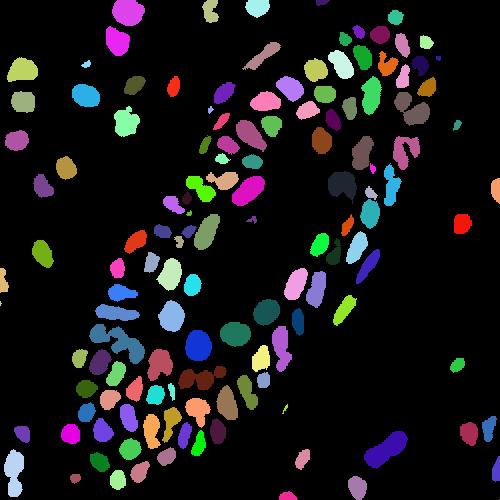} \\
	\begin{minipage}[t]{0.135\textwidth}\centering	(a) image	\end{minipage}
	\begin{minipage}[t]{0.135\textwidth}\centering	(b) true mask \end{minipage}
	\begin{minipage}[t]{0.135\textwidth}\centering	(c) 5\% \end{minipage}
	\begin{minipage}[t]{0.135\textwidth}\centering	(d) 10\% \end{minipage}
	\begin{minipage}[t]{0.135\textwidth}\centering	(e) 25\% \end{minipage}
	\begin{minipage}[t]{0.135\textwidth}\centering	(f) 50\% \end{minipage}
	\begin{minipage}[t]{0.135\textwidth}\centering	(g) gt points \end{minipage}
	\caption{Typical segmentation results using different ratios of points annotation on LC dataset (top row) and MO dataset (bottom row). (a) image, (b) ground-truth full mask, (c)-(f) are results using different ratios of points, (g) the results of using all ground-truth (100\%) points without the detection step. Distinct colors represent different nuclei.} \label{fig:seg:diff-ratio-good}
\end{figure*}

\subsection{Segmentation results using detected points}
The settings, including parameters in the loss function and training details, are the same as those using ground-truth points. The only difference is that compared to the ground-truth points there are errors in the detected points, i.e., false positives, false negatives and localization errors. As a result, the errors in the generated Voronoi labels and cluster labels using detected points increase, which will degrade the performance.

\begin{figure}[t]
	\centering
	\includegraphics[width=0.49\linewidth]{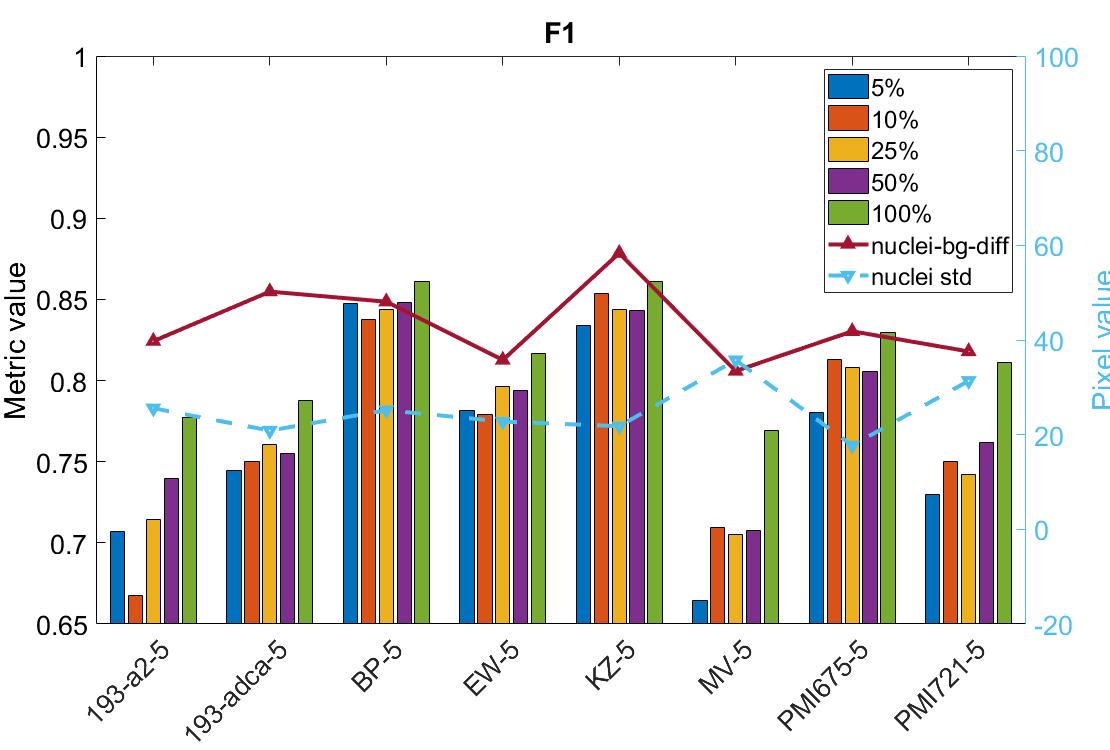}
	\includegraphics[width=0.49\linewidth]{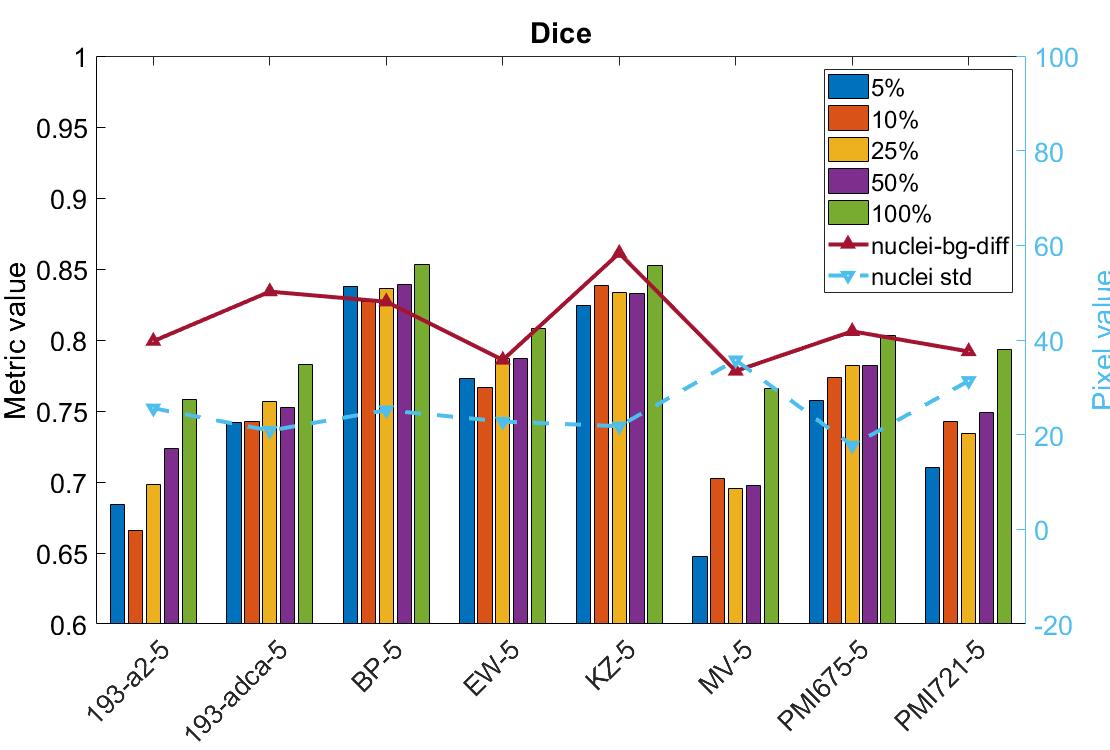}\\
	\includegraphics[width=0.49\linewidth]{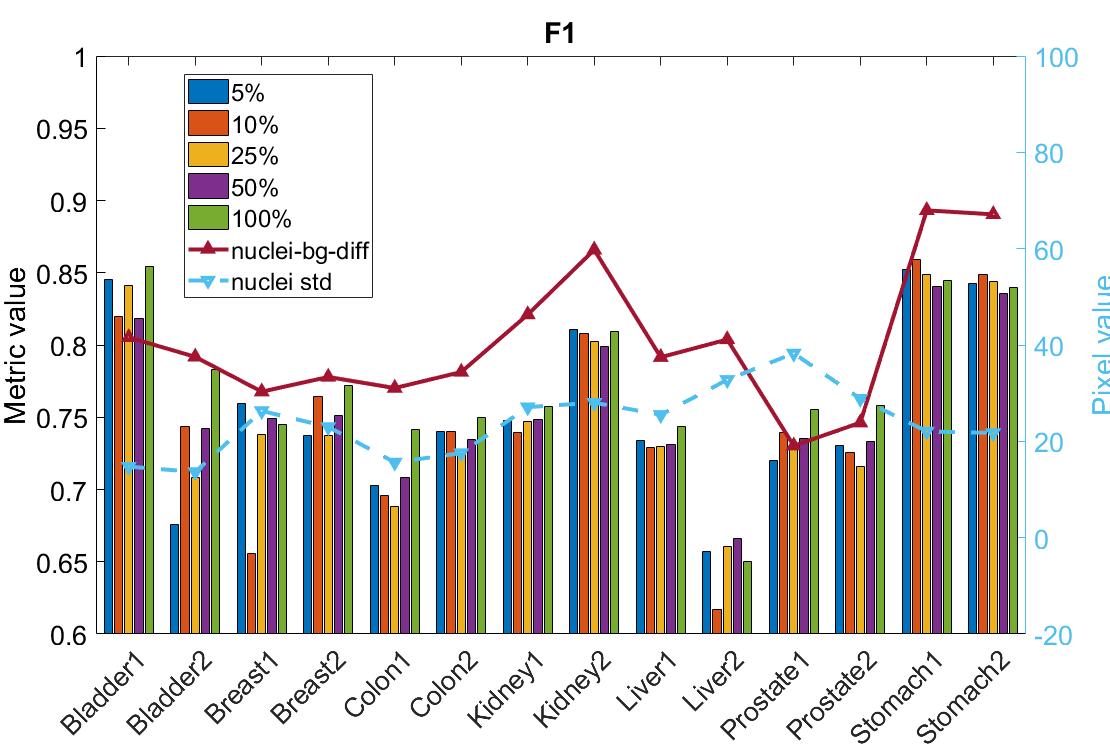}
	\includegraphics[width=0.49\linewidth]{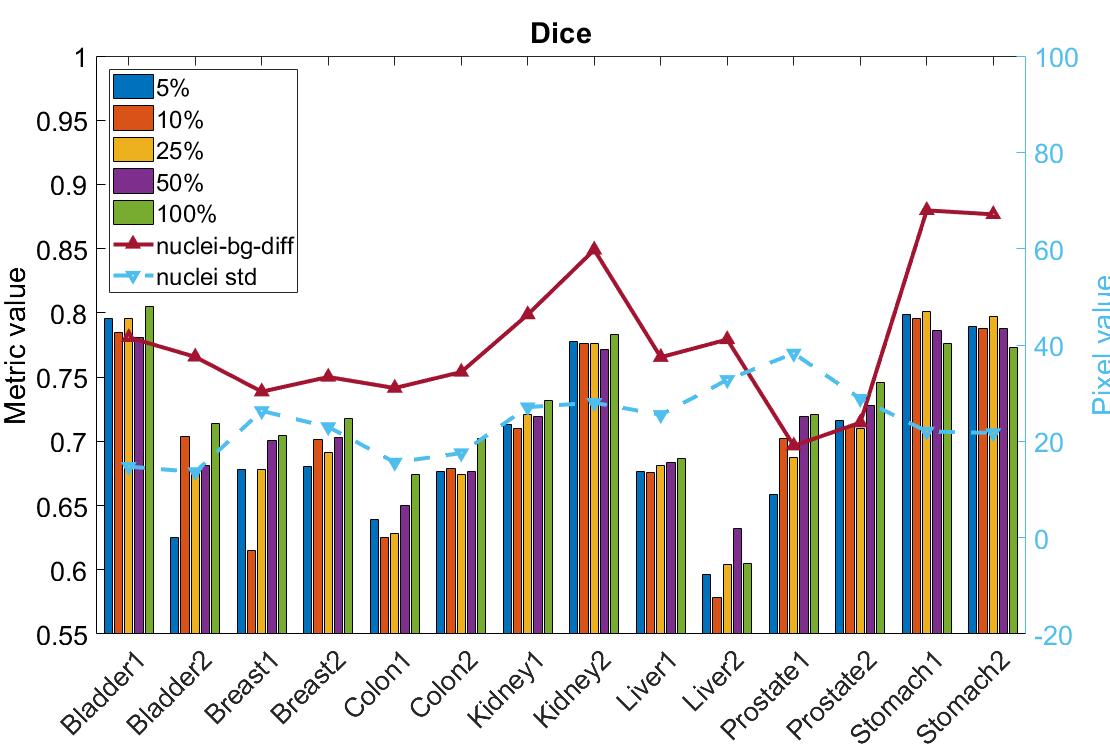}
	\caption{The results of all test images using partial points and ground-truth points (100\%) in the LC dataset (first row) and MO dataset (second row). \textit{nuclei-bg-diff} is the difference between pixel values of nuclei and background. \textit{nuclei-std} is the standard deviation of the pixel values within nuclei.}
	\label{fig:all-img-results-LC}
\end{figure}

\begin{figure*}[t]
	\centering
	\includegraphics[width=0.14\textwidth]{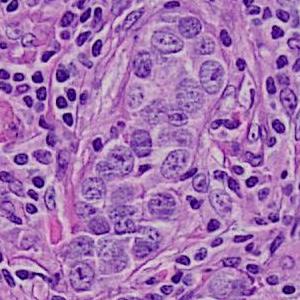} \hspace{-0.08in}
	\includegraphics[width=0.14\textwidth]{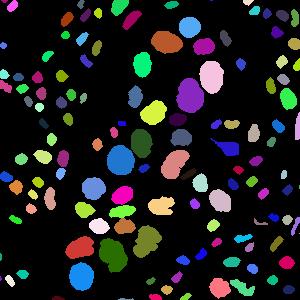}  \hspace{-0.08in}
	\includegraphics[width=0.14\textwidth]{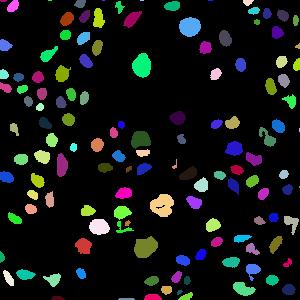}  \hspace{-0.08in}
	\includegraphics[width=0.14\textwidth]{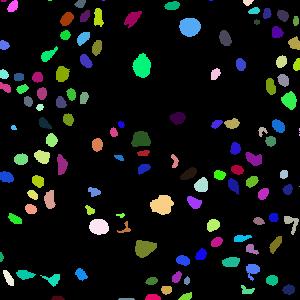}  \hspace{-0.08in}
	\includegraphics[width=0.14\textwidth]{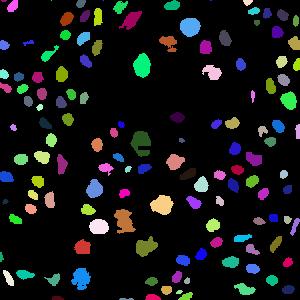}  \hspace{-0.08in}
	\includegraphics[width=0.14\textwidth]{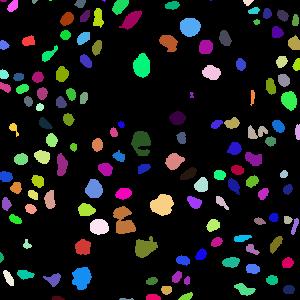}  \hspace{-0.08in}
	\includegraphics[width=0.14\textwidth]{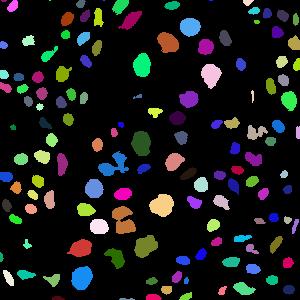} \\	\vspace{0.02in}
	\includegraphics[width=0.14\textwidth]{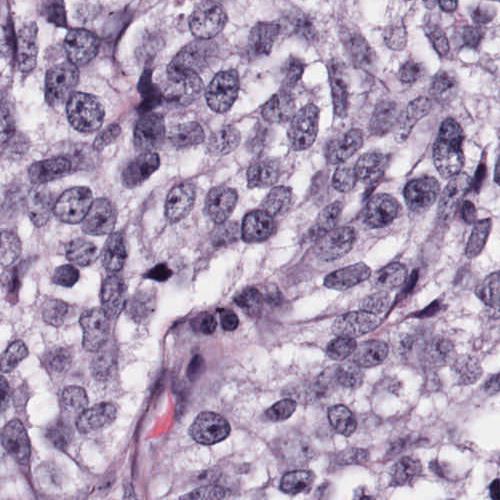} \hspace{-0.08in}
	\includegraphics[width=0.14\textwidth]{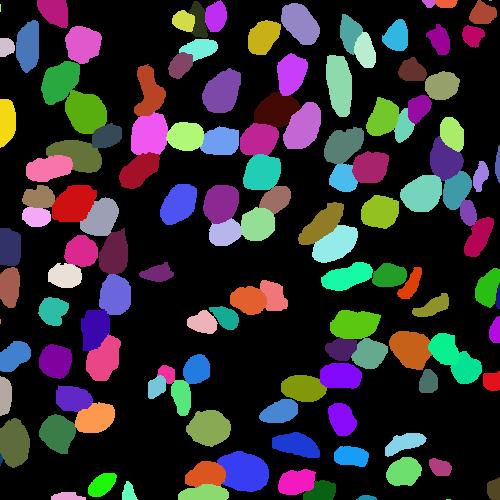}  \hspace{-0.08in}
	\includegraphics[width=0.14\textwidth]{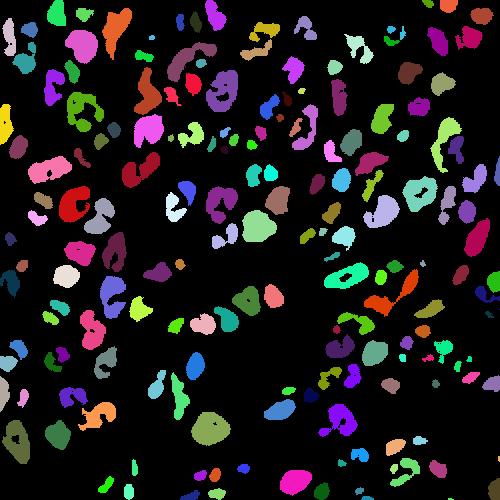}  \hspace{-0.08in}
	\includegraphics[width=0.14\textwidth]{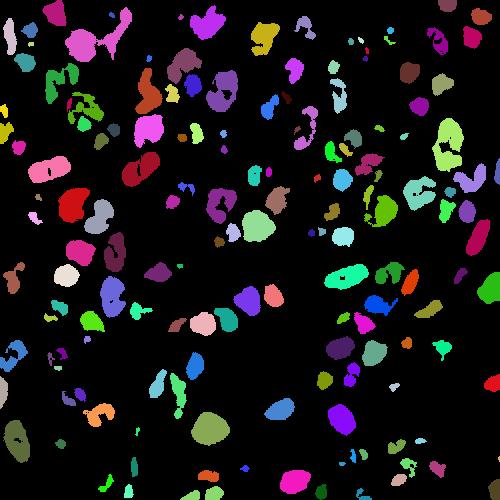}  \hspace{-0.08in}
	\includegraphics[width=0.14\textwidth]{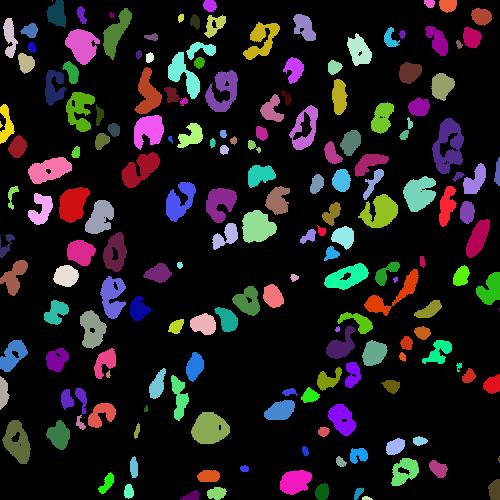}  \hspace{-0.08in}
	\includegraphics[width=0.14\textwidth]{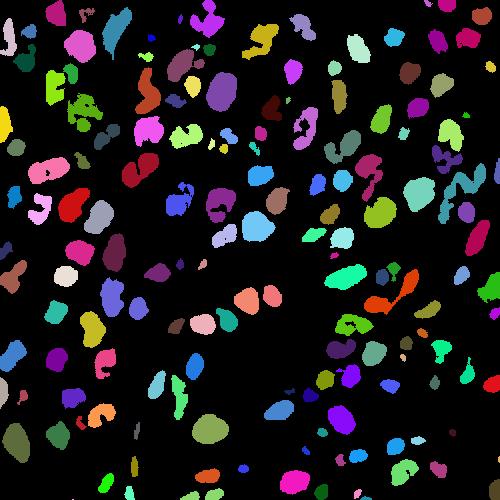}  \hspace{-0.08in}
	\includegraphics[width=0.14\textwidth]{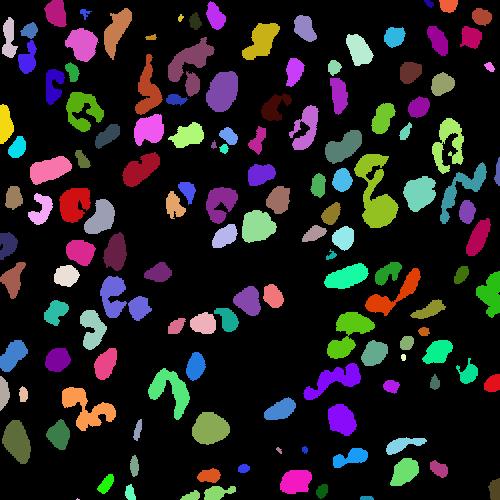} \\
	\begin{minipage}[t]{0.135\textwidth}\centering	(a) image	\end{minipage}
	\begin{minipage}[t]{0.135\textwidth}\centering	(b) true mask \end{minipage}
	\begin{minipage}[t]{0.135\textwidth}\centering	(c) 5\% \end{minipage}
	\begin{minipage}[t]{0.135\textwidth}\centering	(d) 10\% \end{minipage}
	\begin{minipage}[t]{0.135\textwidth}\centering	(e) 25\% \end{minipage}
	\begin{minipage}[t]{0.135\textwidth}\centering	(f) 50\% \end{minipage}
	\begin{minipage}[t]{0.135\textwidth}\centering	(g) gt points \end{minipage}
	\caption{Typical bad segmentation results using different ratios of points annotation on LC dataset (top row) and MO dataset (bottom row). (a) image, (b) ground-truth full mask, (c)-(f) are results using different ratios of points, (g) the results of using all ground-truth (100\%) points without the detection step. Distinct colors represent different nuclei.} \label{fig:seg:diff-ratio-bad}
\end{figure*}

We report the results using detected points from different ratios of initial points annotation in Table~\ref{tab:seg:ratio}. Even with only 5\% annotated points, the proposed framework can achieve satisfactory segmentation performance compared to the fully-supervised ones. As the annotated points increase from 5\% to 50\%, the overall segmentation performance becomes better. This is quite reasonable because the performance is affected by the detection results and the detection error decreases for higher annotated points ratio, as shown in Table~\ref{tab:detect:ratio}. 

To explore the underlying factors that affect the performance, we compute two statistical metrics of the datasets. One is the average difference between the pixel values of nuclei and its surrounding background (\textit{nuclei-bg-diff}) and defined as:
\begin{equation}
\textit{nuclei-bg-diff} = \sum_{i=1}^{N}\frac{A_i}{A_{total}}(\mu_i^{nucleus} - \mu_i^{bg}),
\end{equation}
where $N$ is the number of nuclei in the image, $A_i$ and $A_{total}$ are the areas of the $i$-th nucleus and all nuclei, $\mu_i^{nucleus}$ and $\mu_i^{bg}$ are the average pixel values of the $i$-th nucleus and its surrounding background. We treat the annular area with radius 3 around the nucleus as its background area.
The larger the \textit{nuclei-bg-diff} is, the better the algorithm recognizes each nucleus. The other metric is the standard deviation of pixel values within each nucleus (\textit{nuclei-std}), and defined as:
\begin{equation}
\textit{nuclei-std} = \sum_{i=1}^{N}\frac{A_i}{A_{total}}\sigma_i^{nucleus},
\end{equation}
where $\sigma_i$ is the standard deviation of pixel values within the $i$-th nucleus. The smaller the \textit{nuclei-std} is, the more uniform color in each nucleus, which improves segmentation of the entire nucleus. We show the segmentation results of all test images of LC and MO datasets in Fig.~\ref{fig:all-img-results-LC}, respectively. It can be observed that images with large \textit{nuclei-bg-diff} and small \textit{nuclei-std} have much better segmentation performance, e.g., KZ-5 in LC dataset and Kidney2 in MO dataset. Besides, for those with small \textit{nuclei-bg-diff} and large \textit{nuclei-std}, the performance gap between partial points and ground-truth points are larger than other images, e.g., MV-5 in LC dataset and Prostate1 in MO dataset. Because the nuclei in these images have similar appearance as background pixels and large color variance, thus are hard to be accurately extracted and more sensitive to the errors when using partial points. Typical segmentation results of both datasets are shown in Fig.~\ref{fig:seg:diff-ratio-good} and Fig.~\ref{fig:seg:diff-ratio-bad}.

\begin{table}[t]
	\centering
	\caption{Nuclei detection results on LC and MO datasets using 10 different sets of 10\% initial points.}\label{tab:detect:ablation}
	\begin{tabular}{llccccc}
		\toprule
		Dataset &   & P & R & F1 & $\mu_d$ & $\sigma_d$  \\ \midrule
		\multirow{2}{*}{LC}   
		&  mean  & 0.8544	&  0.9225 	& 0.8868 	& 1.43	& 1.23 \\
		& std	& 0.02057	&  0.02012 & 0.00423  & 0.068 & 0.043 \\ \midrule
		\multirow{2}{*}{MO}   
		& mean	&   0.8227	&  0.8453 	&  0.8330	&  2.92	&  2.11 \\
		& std & 0.03744 	&  0.02148 	 & 0.01041	 & 0.079	 & 0.066 \\
		\bottomrule
	\end{tabular}
\end{table}

\begin{table}[t]
	\centering
	\caption{Nuclei segmentation results on LC and MO datasets using 10 different sets of 10\% initial points.}
	\label{tab:seg:ablation}
	\begin{tabular}{llcccc}
		\toprule		
		\multirow{2}{*}{Dataset} &  & \multicolumn{2}{c}{Pixel-level} & \multicolumn{2}{c}{Object-level}  \\	\cmidrule{3-6}
		& & Acc & F1 & Dice$_{obj}$ & AJI \\ \midrule
		\multirow{2}{*}{LC} 
		& mean  & 0.9278 & 0.7695 & 0.7571 & 0.5880 \\
		& std  & 0.00264 & 0.00381 & 0.00443 & 0.00782 \\ \midrule
		\multirow{2}{*}{MO} 
		& mean  & 0.8982 & 0.7484 & 0.7089 & 0.5158 \\
		& std  & 0.00472 & 0.00753 & 0.00666 & 0.00413 \\
		\bottomrule
	\end{tabular}
\end{table}

\subsection{Sensitivity and generalization analyses}
Two more experiments are conducted to analyze the sensitivity of our method and the generalization performance of the trained models.
\subsubsection{Sensitivity}
To explore how the initial selected points will affect the final performance of our method, we randomly select ten different sets of 10\% initial points, and perform detection and segmentation. The results are reported in Table~\ref{tab:detect:ablation} and Table~\ref{tab:seg:ablation}. The small variances in the metrics indicate that our method is not sensitive to the choice of initial points.

\subsubsection{Generalization}
Generalization performance is an important aspect when applying a method to other datasets. We use the best segmentation model trained on one dataset (LC/MO) to test its performance on the other dataset (MO/LC). The results are shown in Table~\ref{tab:seg:generalization}. When applying the MO models to LC test set (MO $\rightarrow$ LC), they can achieve 92\% to 99\% performance compared to the models trained on the LC dataset. On the MO test set, the models trained on the LC dataset can achieve 95\% to 99\% performance compared to the models trained on the MO dataset. The results illustrate the good generalization performance of our model on different nuclei datasets.

\begin{table}[t]
	\centering
	\caption{Generalization performance of our method on LC and MO datasets.}
	\label{tab:seg:generalization}
	\begin{tabular}{cccccc}
		\toprule		
		\multirow{2}{*}{Train $\rightarrow$ Test} & \multirow{2}{*}{Ratio}  & \multicolumn{2}{c}{Pixel-level} & \multicolumn{2}{c}{Object-level}  \\	\cmidrule{3-6}
		& & Acc & F1 & Dice$_{obj}$ & AJI \\ \midrule
		\multirow{4}{*}{MO $\rightarrow$ LC}
		& 5\%   & 0.9271  & 0.7589 & 0.7418 & 0.5608 \\
		& 10\%  & 0.9213 & 0.7518 & 0.7297 & 0.5555 \\
		& 25\%  & 0.9222 & 0.7551 & 0.7320 & 0.5588 \\
		& 50\%  & 0.9226 & 0.7559 & 0.7336 & 0.5608 \\ \midrule
		\multirow{4}{*}{LC $\rightarrow$ MO} 
		& 5\%   & 0.9004 &	0.7419 &	0.7028 &	0.4884 \\
		& 10\%  & 0.8964 &	0.7338 &	0.6913 &	0.4971  \\
		& 25\%  & 0.8974 &	0.7234 &	0.6886 &	0.4870 \\
		& 50\%  & 0.8970 &	0.7232 &	0.6986 &	0.5030  \\
		\bottomrule
	\end{tabular}
\end{table}

\subsection{Annotation time} Dr. Riedlinger, a board-certified pathologist annotated eight images (one per case) in the LC dataset using points, bounding boxes and full masks, respectively. The average time spent on each image (about 600 nuclei on average) for full masks is 115 minutes, while for bounding boxes 67 minutes. However, it only takes about 14 minutes for all points annotation and less than 2 minutes for 10\% points annotation.

\section{Conclusion}\label{sec:conclusion}

In this paper, we present a new weakly supervised nuclei segmentation method using only a small portion of nuclei locations. In the first stage, a semi-supervised nuclei detection algorithm is proposed to obtain the locations of all nuclei from the partial annotation. In the second stage, we perform nuclei segmentation using the detected points as weak labels. We generate the Voronoi label and cluster label from the detected points and take advantage of the dense CRF loss to refine the trained model. Our method achieves comparable performance as fully supervised methods while requiring much less annotation effort which in turn allows us to analyze large amounts of data.

\balance

\bibliographystyle{IEEEtran}
\bibliography{myref}

\end{document}